\documentclass{sigchi}
\pdfoutput=1
\usepackage{balance}  %
\usepackage{graphics} %
\usepackage{txfonts}
\usepackage{mathptmx}
\usepackage[pdftex]{hyperref}
\usepackage{color}
\usepackage{booktabs}
\usepackage{textcomp}
\usepackage{microtype} %
\usepackage{ccicons}  %
\usepackage[utf8]{inputenc} %

\usepackage{units}

\DeclareMathOperator{\score}{score}
\DeclareMathOperator{\bias}{bias}
\DeclareMathOperator{\reliability}{reliability}

\def\plaintitle{Peer Grading in a Course on Algorithms and Data Structures: Machine Learning Algorithms do not Improve over Simple Baselines}
\def\plainauthor{Authors}
\def\plainauthor{Mehdi S. M. Sajjadi, Morteza Alamgir, Ulrike von Luxburg}
\def\plainkeywords{machine learning; peer grading; peer assessment; peer review; L@S; ordinal analysis; rank aggregation.}

\makeatletter
\def\url@leostyle{%
  \@ifundefined{selectfont}{
    \def\UrlFont{\sf}
  }{
    \def\UrlFont{\small\bf\ttfamily}
  }}
\makeatother
\def\pprw{8.5in}
\def\pprh{11in}

\definecolor{linkColor}{RGB}{6,125,233}
\hypersetup{%
  pdftitle={\plaintitle},
  pdfauthor={\plainauthor},
  pdfkeywords={\plainkeywords},
  bookmarksnumbered,
  pdfstartview={FitH},
  colorlinks,
  citecolor=black,
  filecolor=black,
  linkcolor=black,
  urlcolor=linkColor,
  breaklinks=true,
}

\begin{document}

\toappear{\scriptsize Permission to make digital or hard copies of all or part of this work for personal or classroom use is granted without fee provided that copies are not made or distributed for profit or commercial advantage and that copies bear this notice and the full citation on the first page. Copyrights for components of this work owned by others than ACM must be honored. Abstracting with credit is permitted. To copy otherwise, or republish, to post on servers or to redistribute to lists, requires prior specific permission and/or a fee. Request permissions from permissions@acm.org. \\
{\emph{L@S 2016}}, April 25--26, 2016, Edinburgh, UK. \\
Copyright is held by the owner/author(s). Publication rights licensed to ACM. \\
ACM 978-1-4503-3726-7/16/04\ ...\$15.00.\\
http://dx.doi.org/10.1145/2876034.2876036}
\clubpenalty=10000 
\widowpenalty = 10000

\title{\plaintitle}

\numberofauthors{3}
\author{
 \alignauthor{Mehdi S. M. Sajjadi\\
   \affaddr{Max Planck Institute for Intelligent Systems\thanks{This work was carried out while the authors were still at the University of Hamburg.}}\\
   \affaddr{Spemannstr. 38\\ 72076 Tübingen}\\  
   \email{msajjadi@tuebingen.mpg.de}}\\
 \alignauthor{Morteza Alamgir\\
   \affaddr{University of Hamburg}\\
   \affaddr{Vogt-Kölln-Str. 30\\ 22527 Hamburg}\\
   \email{alamgir@informatik.uni-hamburg.de}}\\
 \alignauthor{Ulrike von Luxburg\\
   \affaddr{University of Tübingen\footnotemark[1]}\\
   \affaddr{Sand 14\\ 72076 Tübingen}\\
   \email{luxburg@informatik.uni-tuebingen.de}}\\
}

\maketitle

\begin{abstract}
  Peer grading is the process of students reviewing each others' work, such as homework submissions, and has lately become a popular mechanism used in massive open online courses (MOOCs). Intrigued by this idea, we used it in a course on algorithms and data structures at the University of Hamburg. Throughout the whole semester, students repeatedly handed in submissions to exercises, which were then evaluated both by teaching assistants and by a peer grading mechanism, yielding a large dataset of teacher and peer grades. We applied different statistical and machine learning methods to aggregate the peer grades in order to come up with accurate final grades for the submissions (supervised and unsupervised, methods based on numeric scores and ordinal rankings). Surprisingly, none of them improves over the baseline of using the mean peer grade as the final grade. We discuss a number of possible explanations for these results and present a thorough analysis of the generated dataset.
\end{abstract}


\section{Introduction}

Peer grading refers to a process where students grade the work of
their co-students based on a scoring rubric
provided by the instructor. While it has been experimented with in the past (see \cite{Topping98}), it has become increasingly popular in the context of MOOCs, where thousands of
students hand in homework that has to be graded (e.g.,
\cite{KulkarniEtal15}). Similar to other crowd-sourcing
scenarios, the hope is that even though students may not be perfect
graders, it might be possible to come up with a \emph{fair} or \emph{accurate} final grade for
each submitted homework by aggregating many such imperfect grades.
The challenge of finding good aggregation
algorithms has been taken up by the machine learning community and a
number of suggestions have been made in the literature (e.g.,
\cite{PiechEtal13,VozHolGil14,RamJoa14,ShahEtal13,Diez13,Walsh14,Venanzi14}).
The bottom line of these papers is that statistical models or machine learning
algorithms are successful at solving this
task.  %

Intrigued by its idea and potential, we used peer grading in an
undergraduate course in computer science at the University of Hamburg:
the course on algorithms and data structures (AD).
Throughout one semester, students solved 6 exercise 
sheets with 2-5 exercises each. 6 Teaching
assistants (TAs) graded all submissions in the traditional way, but at the same time we
applied peer grading on the same set of submissions. The data generated during
the whole semester has been made publicly available, the link can be
found on our homepages. We refer to it as \emph{AD data} below. 
At the end of the semester, we applied various algorithms to aggregate
the peer grades and to analyze our data. Contrary to other
researchers we found that none of these methods show a satisfactory
improvement over a simple baseline. We will discuss a number of
possible explanations for these results and analyze sources of problems that
prevent more sophisticated methods to succeed.

\section{Our setup}

For our peer grading experiment we selected the course on algorithms
and data structures at the Department of Computer Science, University of Hamburg in the winter term 2014/15.
The course is compulsory for all
bachelor students in computer science and had 219 active participants
in total. We had 6 TAs grade all homework in the traditional way. Roughly once every two weeks, students received an exercise sheet as homework, resulting in 6 sheets with 19 individual exercises in total. The exercises consisted of puzzles on algorithms or proofs about
algorithmic properties. For example, the students were asked to apply a given algorithm by hand on a small input or they were asked to design an algorithm to solve a particular problem.
In each exercise, students could collect a certain number of points.
Getting at least 50\% of all homework points in the semester was a necessary requirement to pass the course. Exercises were solved and submitted by
groups of up to three students. This semester we had a total of 79 groups. For each individual exercise, every group handed in their 
submission via an online system (we used a modification of
the moodle open source platform\footnote{https://moodle.org/}).
After the submission deadline, a sample solution together with
assessment criteria was posted online by the course instructor. We tried
to describe all possible correct solutions to the
exercises, point out pitfalls or common errors and discuss details
about the grading procedure. Grading took place simultaneously in
the following three different ways.

\textbf{(i) Self grading.} The students were asked to read and
understand the sample solution and then grade their own submission first. While they solved exercises in groups of three, the self grading took place individually, resulting in 3 self grades per submission.

\textbf{(ii) Peer grading.} Each student graded 2 randomly selected submissions of other groups, resulting in about 6 peer grades per submission. Grading was double-blind.

\textbf{(iii) TA grading.} All submissions were furthermore randomly assigned to and graded by one of the 6 TAs.

In two exercise sheets, we experimented with slightly altered settings. In one case, we increased the number of peer grades by each reviewer from 2 to 5, which results in about 15 peer grades for each submission. On another exercise sheet, we applied ordinal peer grading: Instead of assigning every submission an absolute grade, we gave each student a total of 5 peer submissions and asked them to only rank these submissions from worst to best.

In order to ensure that the students take their peer grading duties
seriously, we made a reasonable peer grading performance a mandatory
requirement to pass the course: students were required to participate
in the peer grading of at least 5 exercise sheets,
and we announced that sloppy grading behavior would not be tolerated.
Moreover, the peer grades contributed to the final grades for the submissions
with a weight of 20\% (with the other 80\% being the respective TA grade). An anonymous questionnaire at the end of the semester revealed that about half 
the students liked the concept of peer grading, had the impression that it helped 
them deepen their understanding, and would like to repeat it in another
course. The other half of the students did not like it, mainly because
it had taken too much of their time.

\begin{figure*}[tb]
\includegraphics[width=0.49\textwidth]{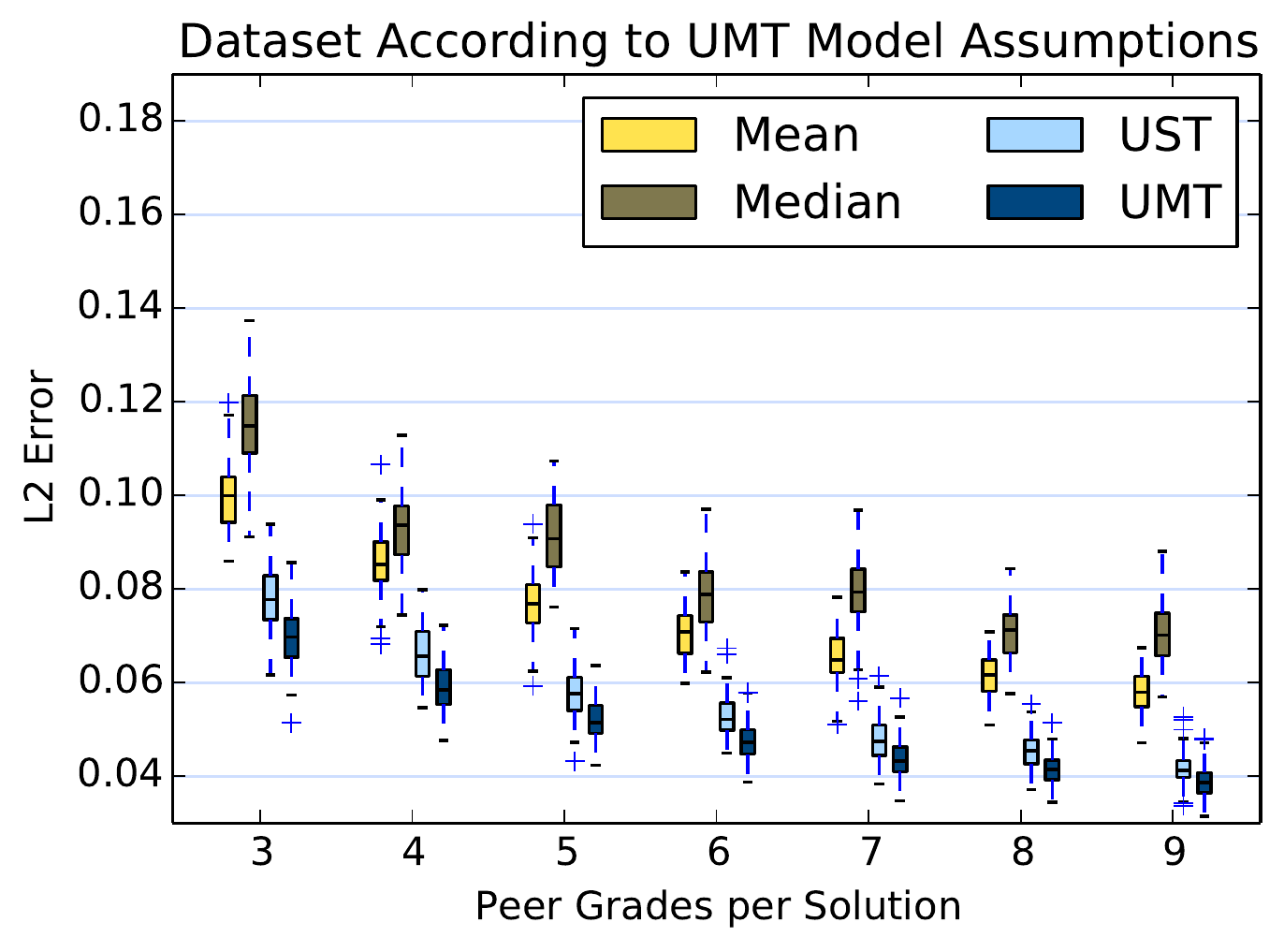}
\hfill 
\includegraphics[width=0.49\textwidth]{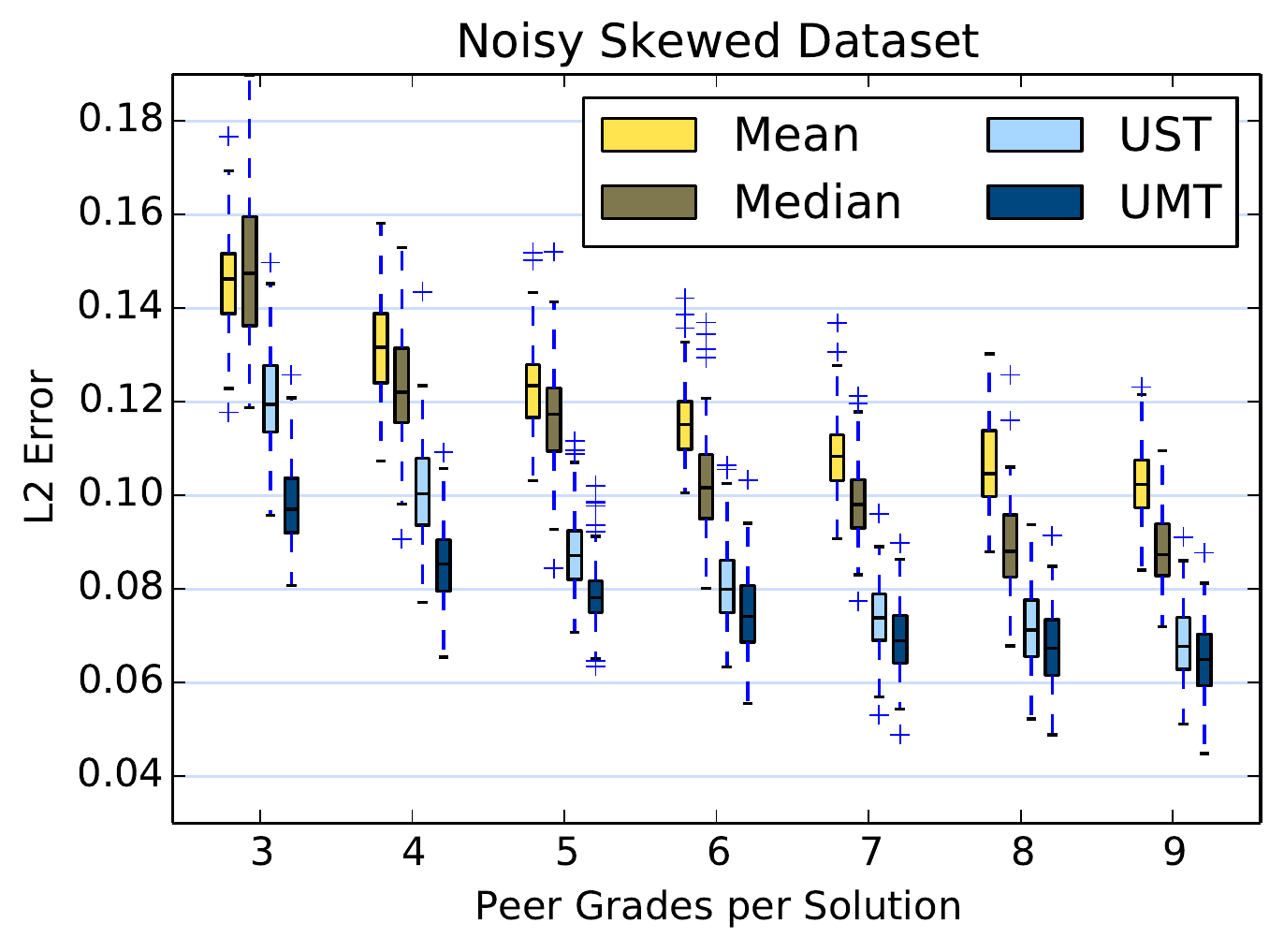}\\
\includegraphics[width=0.49\textwidth]{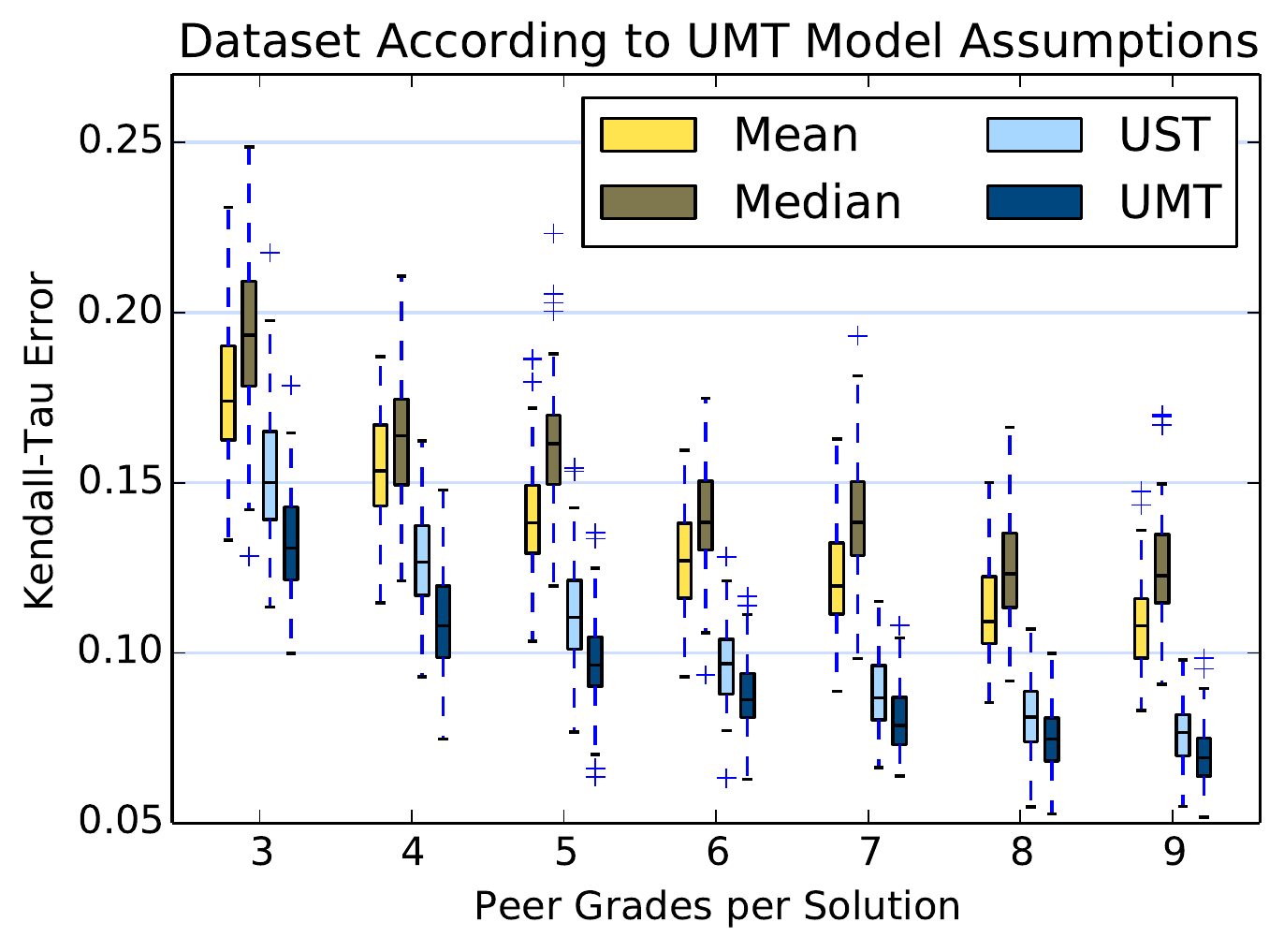}
\hfill 
\includegraphics[width=0.49\textwidth]{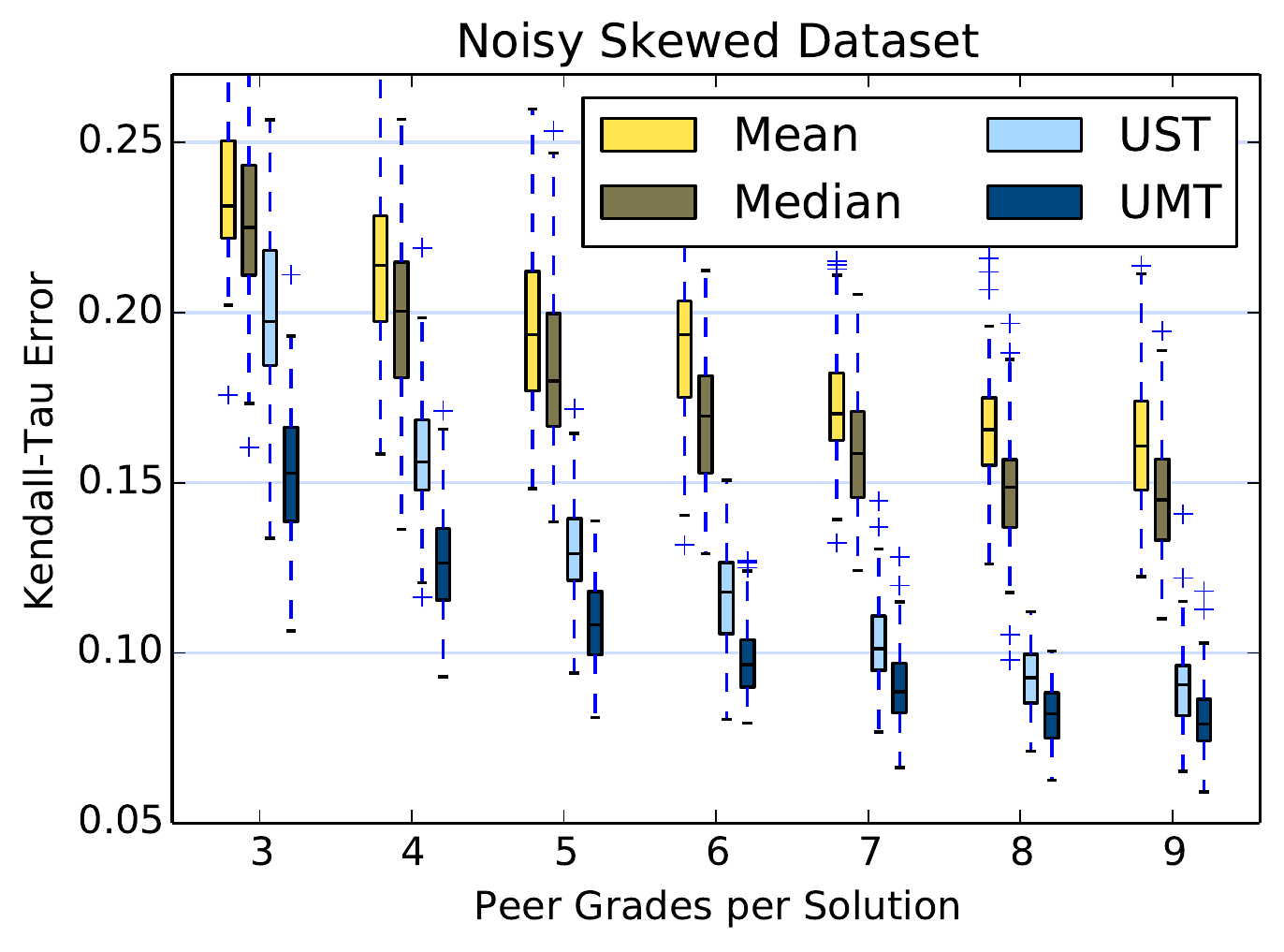}
\caption{
Fitting UST and UMT on artificial data (100 submissions, 100 graders,
5 exercises).
The panels show the $L_2$ and Kendall-$\tau$ errors of the estimated grades with respect to the actual underlying true grades. The box plots show the distribution over the errors of 100 independently generated datasets while the number of peer grades per submission increases along the x-axis.
Left: Dataset generated according to UMT's model assumptions with reasonably realistic hyperparameters (true scores generated with $\mathcal N(\nicefrac{1}{2}, \nicefrac{1}{6^2})$, bias $\mathcal N(0, \nicefrac{1}{8^2})$ and reliability $\Gamma(3, \nicefrac{1}{30})$).
Right: Skewed true score distribution according to $\text{Weibull}(\nicefrac{3}{2}, \nicefrac{1}{3})$ and additionally, 20 graders always draw scores uniformly at random while bias and reliability for the remaining 80 graders is modeled as before.
}
\label{fig-artificial}
\end{figure*}

\section{Algorithms for estimating true grades}

From a statistical or machine learning point of view, the most interesting questions
in the context of peer grading are the following.

{\bf (i) Unsupervised setting.} In the
  complete absence of TA or instructor grades, is it possible to
  take several ``imperfect'' peer grades and to aggregate them into a 
  ``fair'' or ``accurate'' final grade for each submission?
  MOOCs traditionally have this setting.

{\bf (ii) Supervised setting} In a setting where partial grading is available by an
  instructor, is it possible to predict or recover the instructor's grades based
  on aggregated peer grades? If instead of a single instructor, partial grading
  is performed by several TAs, is it possible to predict missing
  grades in the quality of TA grades or even better?
  This is a realistic scenario in large university courses.

{\bf (iii) Adversarial setting} Is it possible to come up with a grading scheme that is robust
  against adversarial behavior of individuals (such as students deliberately giving everyone very high grades or trying to downgrade others) or adversarial attacks of groups of students (such as a cartel of students that act together in order to fool the grading procedure)?

Our work focuses on the first two questions. Let us introduce some
notation. Over the course of the semester, students solved and
graded several exercises. Fix one exercise and suppose
that submissions $s_1, s_2, \ldots$ to this exercise have been handed in
by the students.
Consider a set of graders $g_1, g_2, \ldots$. By
$\score(s,g)$
we denote the score given to submission $s$ by grader
$g$. In the unsupervised scenario, we do not have any data on what the
``true'' score of each submission should be.
Instead, the standard approach is to simply {\em define} the true score $\score(s, true)$ as the population
average over the scores of {\em all} potential graders. As we only
observe the scores of few graders in practice, the goal of
an unsupervised algorithm is to correct for inaccuracies that
are introduced by the actual grading procedure.
In the supervised setting the true score may be given by the instructor or TAs. The goal of supervised peer grading algorithms is to give an estimate $\score(s,estimated)$ of the true scores for all submissions based on the set of
the given peer grades.

Depending on whether the focus is on the absolute values of the scores
or just on the ranking of the submissions, the model performance is evaluated using different error functions. We use the $L_2$ error to compare the absolute score values
and the Kendall-$\tau$ error to compare the rankings induced by the
scores. The Kendall-$\tau$ error between two rankings counts the
number of pairs of items for which the ranking order is inversed. For each pair of items an error of 0 is given for agreement, 1 for inversion, and an error of 0.5 if
exactly one of the rankings gives both items the same rank. The final
Kendall-$\tau$ error is the mean over the errors for all possible pairs.

\begin{figure*}[tb]
\includegraphics[width=0.49\textwidth]{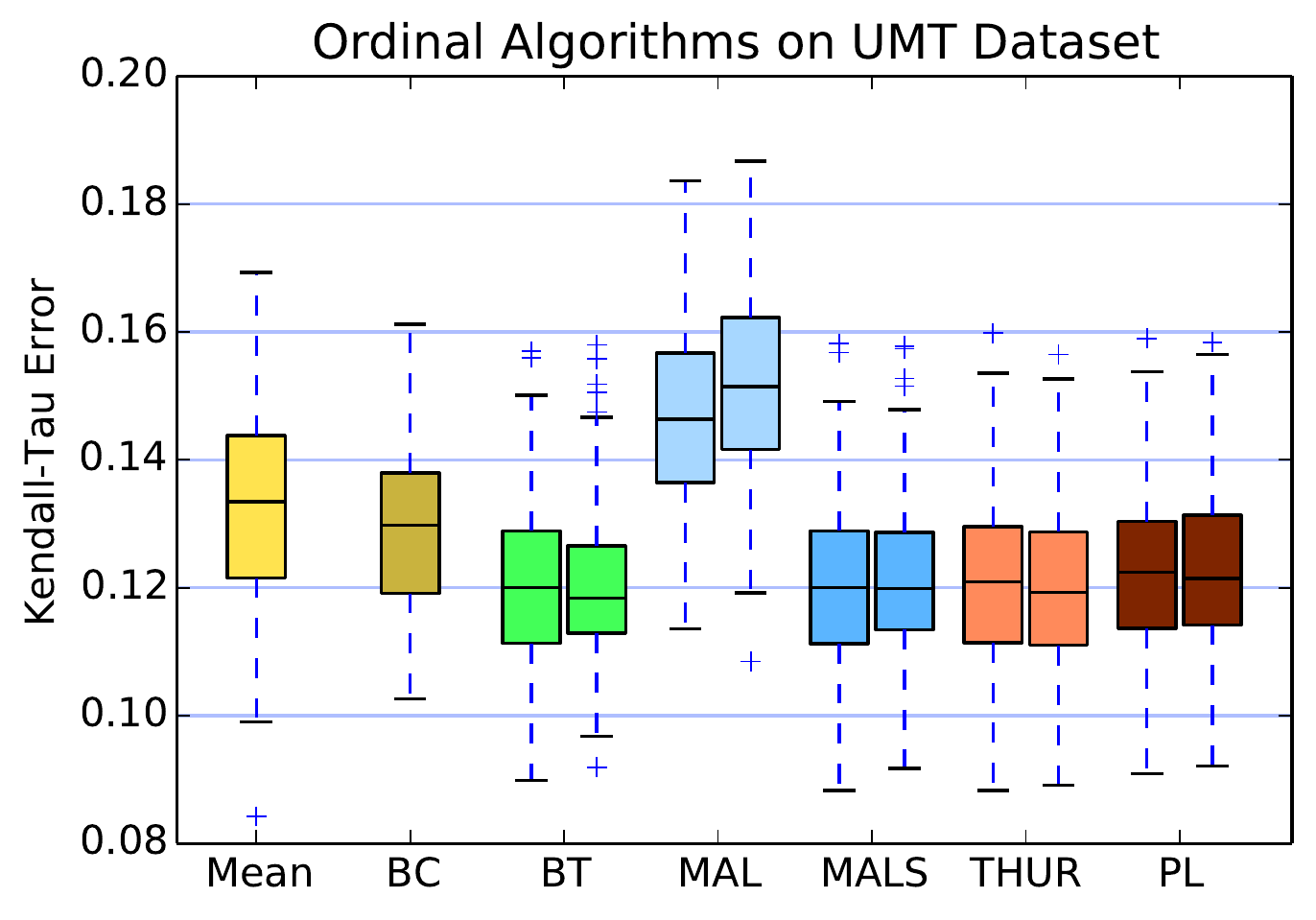}
\hfill 
\includegraphics[width=0.49\textwidth]{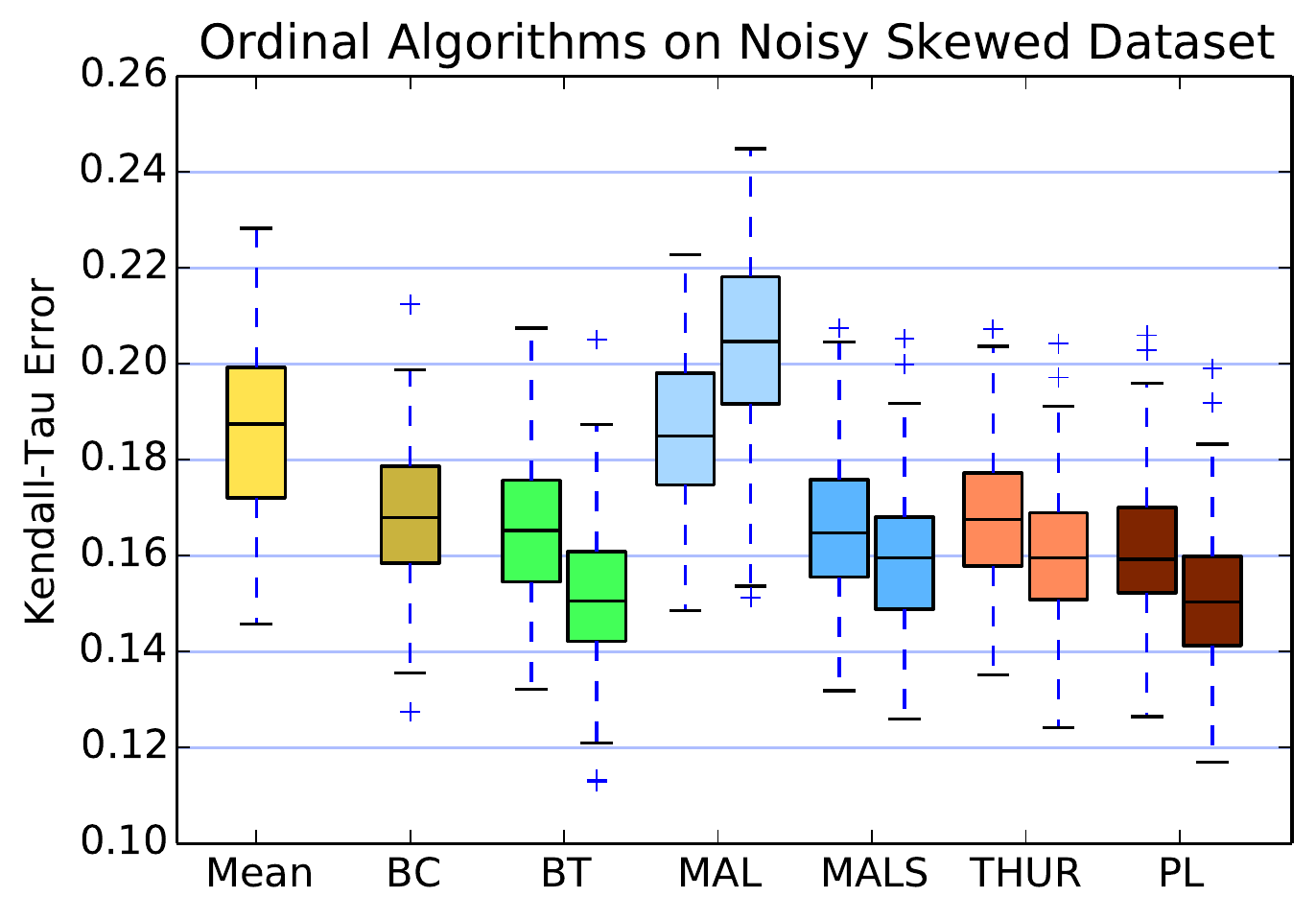}
\caption{Fitting the ordinal algorithms on artificial data (100 submissions, 100 graders,
1 exercise, 6 peer grades per submission) according to UMT model assumptions (left) and on the noisy skewed dataset (right) as in Figure~\ref{fig-artificial}. The two bars for BC, MAL, MALS, THUR and PL show the performance without (left bar) and with (right bar) reliability estimation.}
\label{fig-artificial-ordinal}
\end{figure*}

\subsection{Unsupervised models}
{\bf Mean} and {\bf Median}. We simply estimate the true score of a submission by taking the mean (resp. median) of all peer grades for this submission. These algorithms serve as baseline.

{\bf Unsupervised-single-task (UST)}. Following \cite{PiechEtal13},
  we make some model-based assumptions. Fix
  a single exercise. We assume that
  the true scores of all submissions to this exercise are normally
  distributed as $\mathcal N(\mu_{score}, \sigma_{score}^2)$.
  Each grader $g$ has an inherent $\bias(g)$ and a certain $\reliability(g)$.
  The intuition is that the bias models the tendency of a grader to generally give high scores or to be very strict, whereas the reliability accounts for
the variance in the grading performance. Formally, if a submission $s$ has a true  $\score(s,true)$, then the grade reported by $g$ is
  normally distributed as 
\begin{align*}
\mathcal N\Big(\score(s,true) + \bias(g), 1/\reliability(g)\Big). 
\end{align*}
As the actual true score is not known, it is replaced by the current estimate $\score(s,estimated)$ in practice. The bias of all graders is distributed as 
$\mathcal N(0, \sigma_{bias}^2)$ and the reliability is Gamma-distributed as 
$
\Gamma(\alpha, \beta).
$
The hyperparameters $\mu_{score}$, $\sigma_{score}$, $\sigma_{bias}$, $\alpha$,
$\beta$ can be used to control the weight of bias and reliability estimation as 
well as the strength of regularization.
The goal is to fit the model to the observed data and to learn 
the parameters $\bias(g)$, $\reliability(g)$ and $\score(s,estimated)$. For this purpose we use the EM algorithm, which \cite{PiechEtal13} found to give results similar to those obtained by more elaborate Gibbs sampling procedures.

{\bf Unsupervised-multiple-tasks (UMT)}.
In our course, each student graded a total of about 57 different submissions for 19 exercises over the whole semester. While the UST model just learns from one exercise at a time, the UMT model learns the parameters jointly over all exercises. UMT's basic assumption that the bias and reliability of a grader are inherent attributes and do not vary over different exercises could lead to more accurate estimates as there is much more data to work with.

Evaluating UST and UMT on artificial data confirms the finding in \cite{PiechEtal13} that such models do a remarkable job at recovering the true underlying score, see Figure~\ref{fig-artificial} for an illustration. As more grades per submission are 
added, UST performs almost as well as UMT and they both beat simple algorithms such 
as mean and median by far. For example, UMT only needs 4 grades per submission to reach the error rates of the mean algorithm with 9 peer grades as seen in the left figures. In particular, the models even work reasonably well on artificial data with a model mismatch, i.e.\ data that has not been generated according to the model assumptions.

{\bf Ordinal models}. It has been suggested
  that instead of collecting numeric grades for each
  submission (cardinal peer grading), it might be more reliable to ask each student to only \emph{rank} a set of submissions. Using those partial rankings of each reviewer, the purpose of the ordinal algorithms is to generate an overall ranking of all submissions and to assign them cardinal rankings following a certain score distribution.
  
  A very simple algorithm for this purpose is \textbf{Borda Count (BC)}. Given a ranked set of submissions $s_0<\ldots<s_k$ by a reviewer (where $k+1$ is the number of reported grades by each reviewer), the algorithm gives submission $s_i$ the score $i$ that is the number of submissions that were ranked lower than $s_i$. The sums of these scores for each submission determine the overall BC score. A number of more elaborate models have been described for this purpose in \cite{RamJoa14}: \textbf{Mallows (MAL)} and \textbf{scored Mallows (MALS)}, \textbf{Bradley-Terry (BT)}, \textbf{Thurstone (THUR)} and \textbf{Plackett-Luce (PL)}.
  
  We will focus here on the BT model, which assumes that the likelihood of switching the ranking of a pair of submissions depends on the distance of their true scores and on the reliability of the grader:
  \begin{gather*}
	P(\score(s_1,true)>\score(s_2,true)) \\
	= \frac{1}{1+\exp\left(-\reliability(g)\cdot(\score(s_1,g)-\score(s_2,g))\right)}
  \end{gather*}
  The scores and reliabilities are then estimated using an alternating stochastic gradient descent algorithm.  The performance of the ordinal algorithms in comparison to the Mean baseline on artificial datasets is shown in Figure~\ref{fig-artificial-ordinal}. Note that the Mean baseline is computed on the actual grades as opposed to the ordinal models that only have access to the rankings. Nonetheless, all ordinal models but MAL beat the Mean algorithm. The reliability estimation only shows improvements in the right figure, which is unsurprising, as ordinal grading implicitly applies a threshold for inaccuracies in the reported grades: while it is likely that a reviewer misses the true cardinal grade, getting just the order right is much easier, so the effects of the reliability estimation are only visible here when a large portion of the reviewers is reporting random grades. Increasing the magnitude of the bias values for the reviewers decreases the performance of the Mean algorithm while that only has a negligible effect on the ordinal models (not shown in the figure).
  
  \begin{figure*}[tb]
\centering
\includegraphics[width=0.63\textwidth]{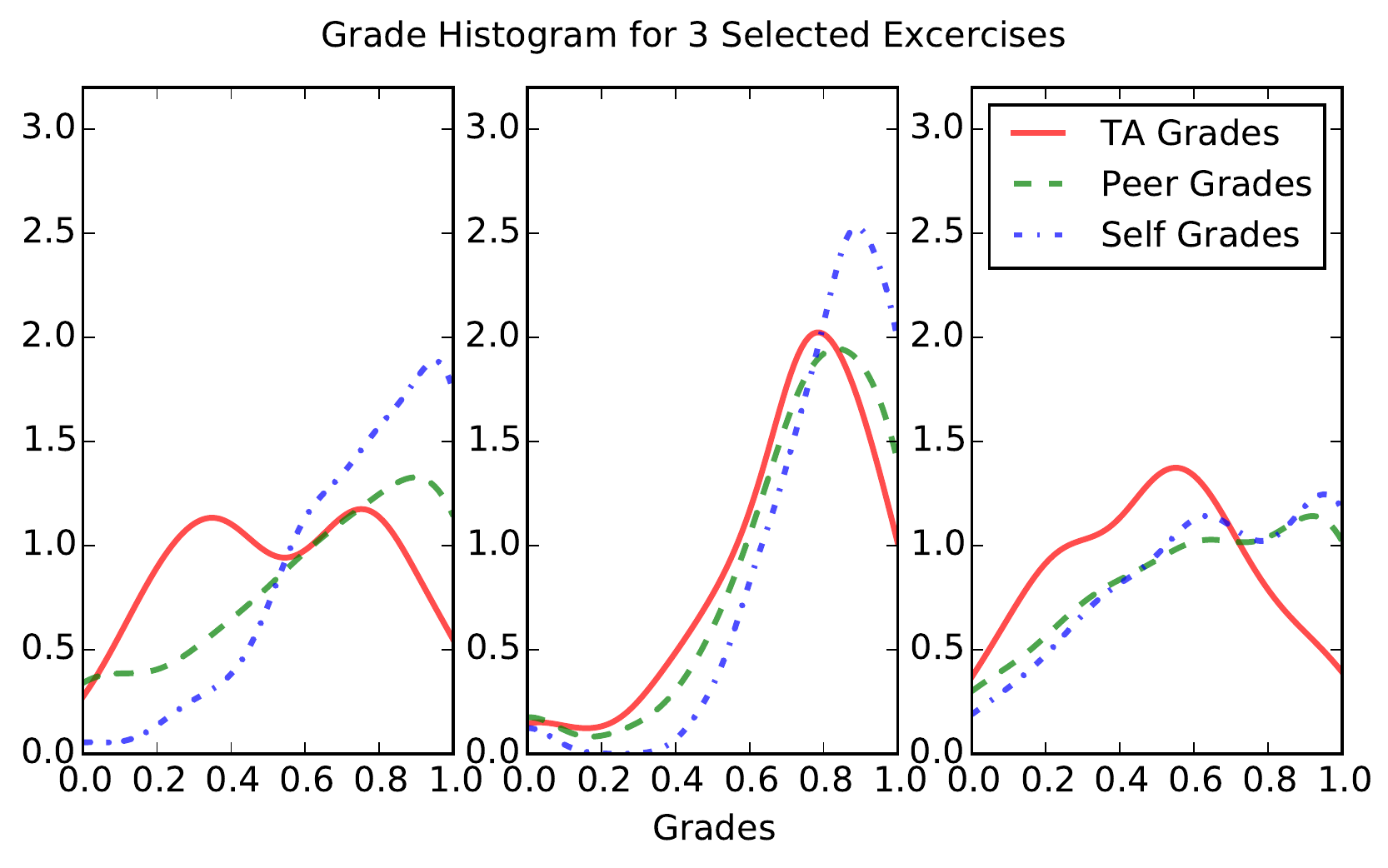}
\hfill
\includegraphics[width=0.363\textwidth]{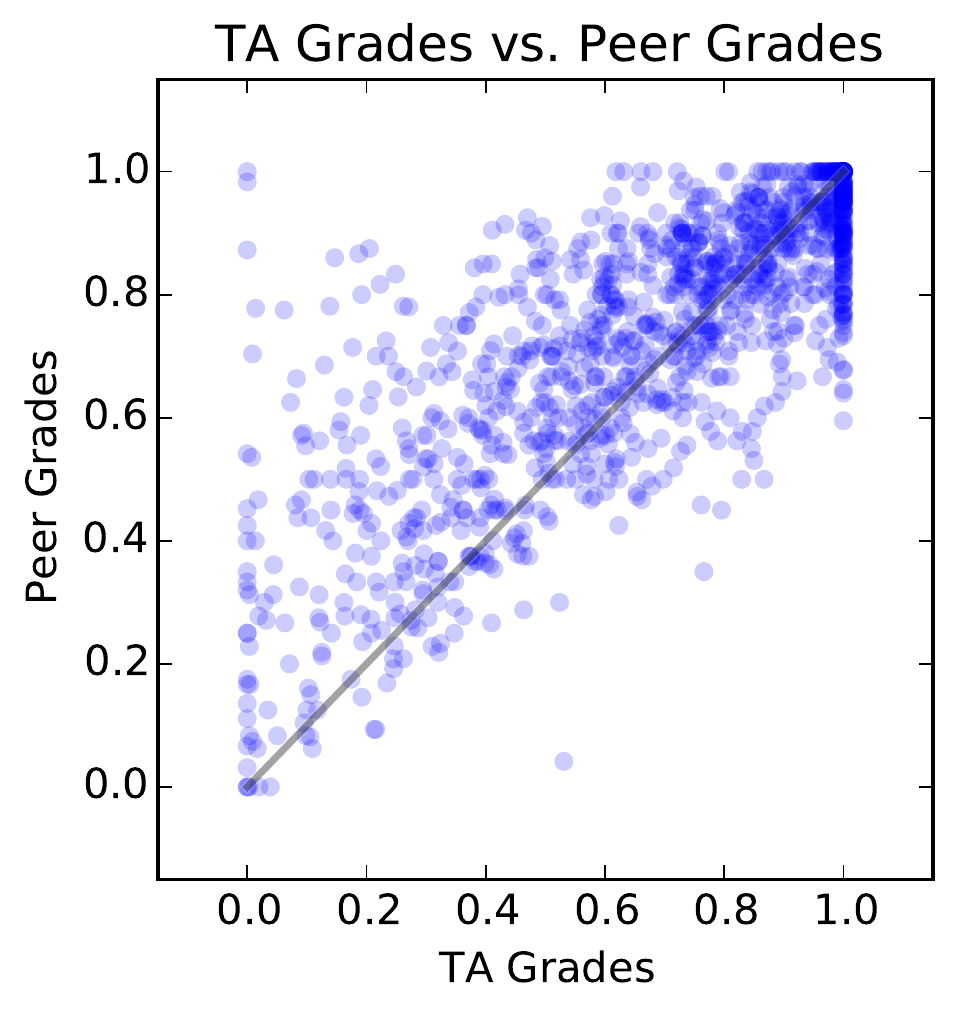}
\caption{TA grades compared to mean peer grades. Left: Smoothed and rescaled score 
histograms for three selected exercises. Right: Overview of all submissions and 
their grades. Each circle depicts one submission with its position indicating the 
TA score and mean peer grade (jitter added to avoid clustering due to different 
scales). If the TAs and students fully agreed on the scores, all circles would lie 
on the indicated diagonal.}
\label{fig-histogram-self-peer-ta}
\end{figure*}

\subsection{Supervised models}
In the supervised models, the goal is to learn to use the peer grades
to predict the true grades as given by an instructor. In our case, we
take the grades provided by the TAs as ground truth. We consider the
following two approaches. 

{\bf Supervised-naive (SN)}. As baseline we use the following
  naive algorithm. For all exercises, the submissions are split into a training set and a test set. We use the TA grades as ground truth for the training sets to estimate the overall student grader biases. With those bias estimates, we compute the bias-corrected mean scores. Denote the submissions that student $g$ graded over the whole course in the training set by $s_1,\ldots,s_n$. Taking the TA grades as true scores, we calculate the overall bias for the student as
  \[ \bias(g) = \frac{1}{n} \sum_{i=1}^n \score(s_i,g)-\score(s_i,true). \]
  On the test set, let $g_1,\ldots,g_k$ be all students who graded submission $s$. The bias-corrected mean score is then given as
  \[ \score(s,estimated) = \frac{1}{k} \sum_{j=1}^k \score(s,g_j) - \bias(g_j). \]

{\bf Supervised-multiple-tasks (SMT)}. Another approach is to
  incorporate the TA grades directly into the UMT model. For some
  subset of the submissions, the respective TA grades are put into the
  model like any other peer grade, but the TA reliabilities are set to
  a high constant. This automatically corrects the student reviewer biases towards
  the TA grades and also gives higher reliabilities to students that
  grade similar to the TAs. To see how much this improves the overall accuracy, the
  error is only computed on submissions whose TA grades are unknown to the model.

\subsection{Linking the data}
While all introduced models estimate the scores, bias and reliability in a very direct way, more accurate results might be obtained by linking up the data in other ways. For example, it stands to reason that students with higher grades on their own homework submissions are likely to be more reliable at grading than students with lower grades. One may even take it a step further by making the homework scores dependent on their grading reliability in order to motivate them to put more effort into the peer grading. Another example is that better performing students might have higher standards when grading other submissions, effectively resulting in a negative bias.

Assuming the data shows strong correlations in this regard, there are several ways to incorporate these additional ties into the model. A very simple approach is fitting a linear function between the own homework scores to the bias/reliability and to use that in the computation of the final grades. Using UMT, one can also build a hybrid model by scaling the estimated reliability with the output of the linear function that takes the own score as input. Such approaches have been shown to improve the model accuracy, see e.g.\ \cite{PiechEtal13, MiYeung15}. We will study these correlations in our dataset and analyze whether they can be used to improve the score estimates.

\section{Our dataset and its analysis}
\begin{figure*}[tb]
\includegraphics[width=0.5\textwidth]{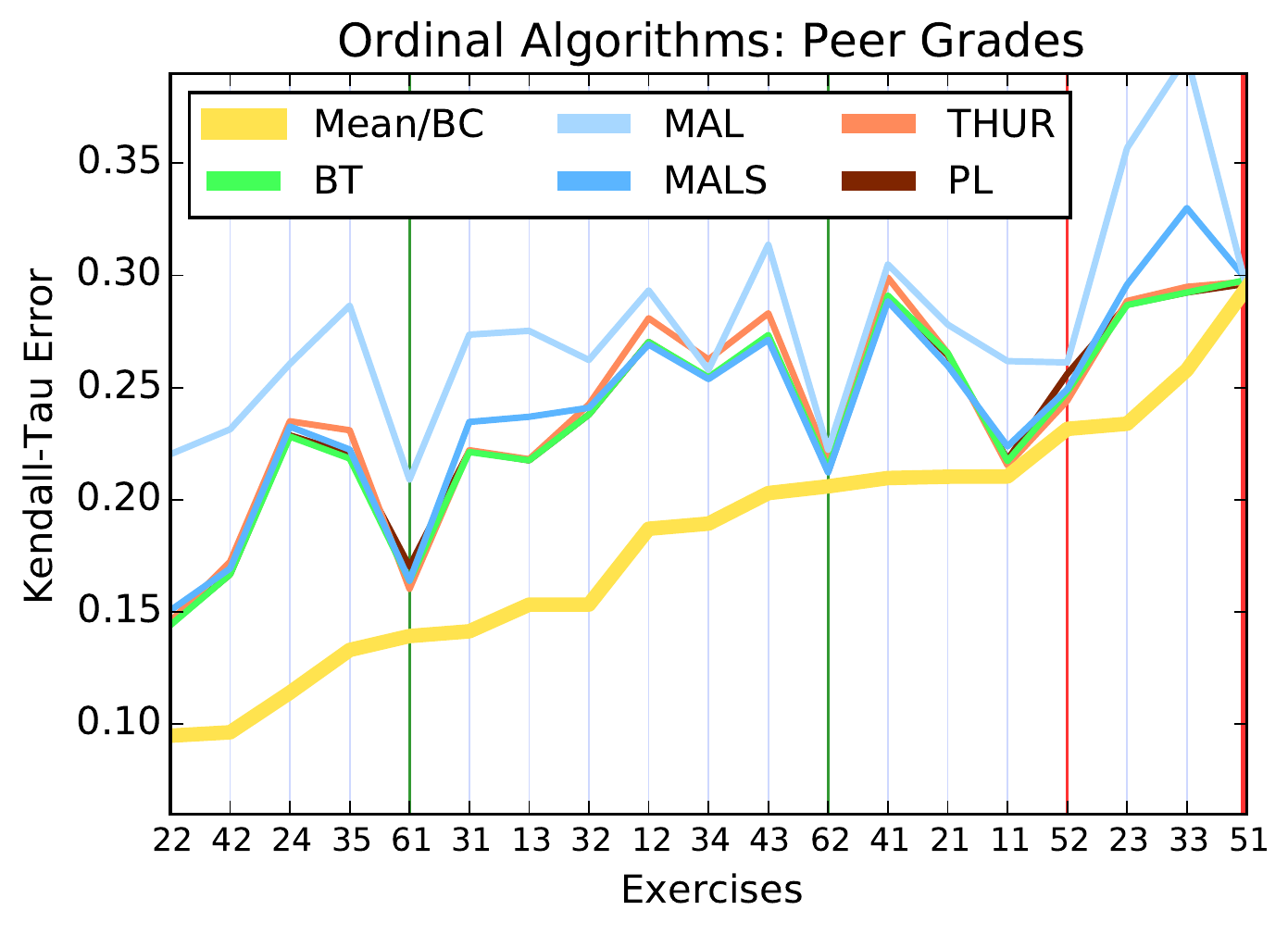}
\includegraphics[width=0.5\textwidth]{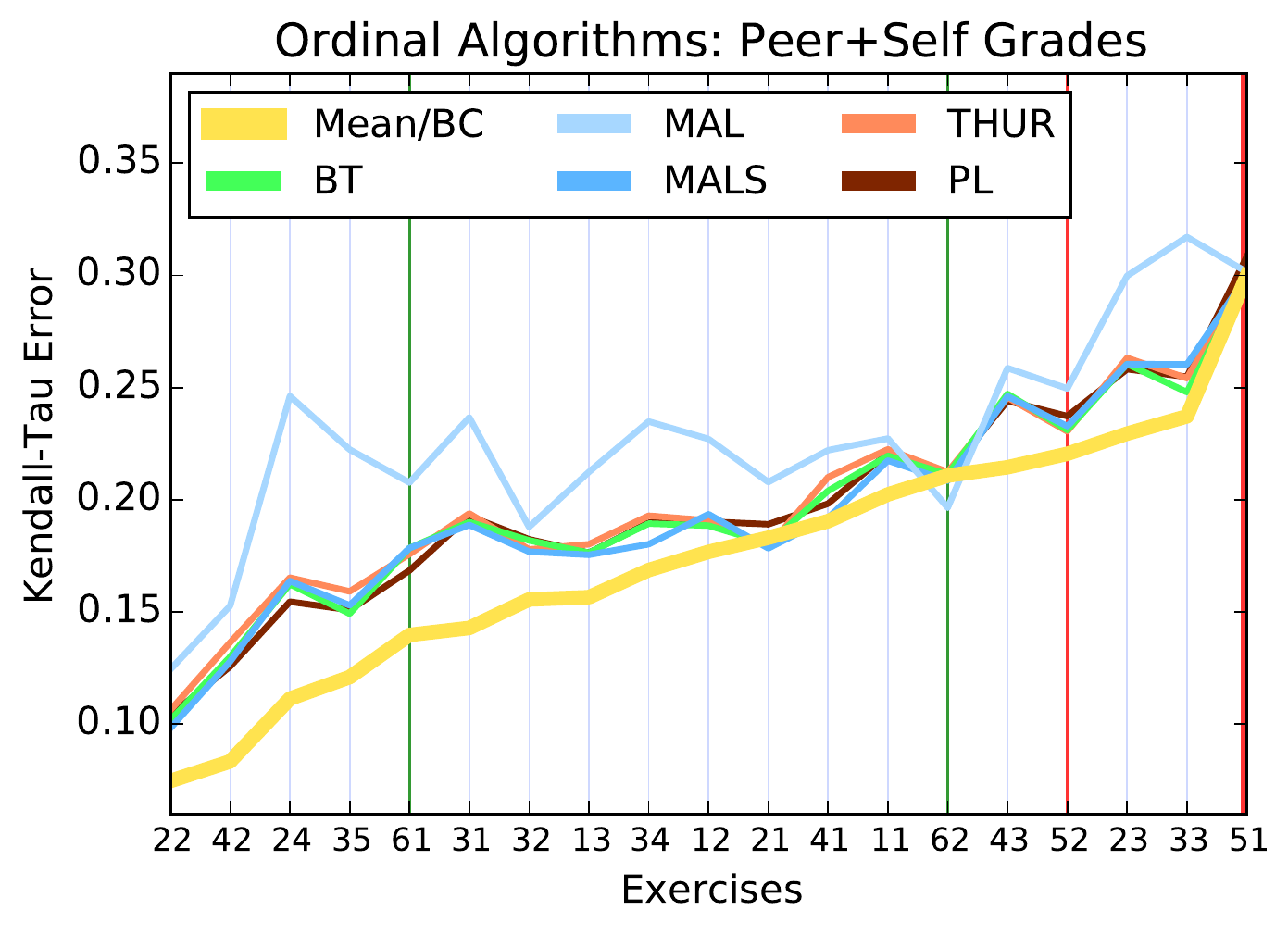}
\caption{Comparing the performance of the ordinal models without reliability estimation on the AD dataset with the mean algorithm as baseline. The exercises on the x-axis are sorted by the error of the mean algorithm. In exercises 61 and 62 (green), each student reported 5 instead of 2 peer grades. Exercises 51 and 52 (red) were graded ordinally, so the BC algorithm is used as a baseline instead of the mean.}
\label{fig-ordinal-results}
\end{figure*}

\subsection{AD data: first observations}
We will begin with an overview over the raw data. To be able to compare performances across different exercises, we rescale the scores of each exercise to lie in the interval $[0,1]$. A different standardization of shifting and rescaling the scores to have mean 0 and variance 1 (z-scores) leads to very similar results.
To get a first impression of our data, consider the plots on the left in
Figure~\ref{fig-histogram-self-peer-ta}. In each panel, the figure
shows histograms of scores for a particular exercise. These already
show a number of interesting points.  Not very surprisingly, we see
that self grades are often higher than peer grades, which in
turn tend to be higher than the grades given by the TAs. 
For some of the exercises, it looks like the peer grade histograms are 
``shifted versions'' of the TA histograms but this is not always the case.
To the contrary, sometimes the overall
characteristics and shapes of the histograms are quite different. 
In general the scores do not seem to be normally distributed. 
The first obvious reason is that the scores are bounded to a fixed interval which can lead to several artifacts. The score distributions are often skewed,
 for example when the exercise was easy and many students
  got full marks for their submission. In many cases the score distribution is
  clearly bi- or multi-modal: this can arise if a large number of students solved only part of the exercise whereas others solved it completely.

\begin{figure*}[tb]
\includegraphics[width=0.49\textwidth]{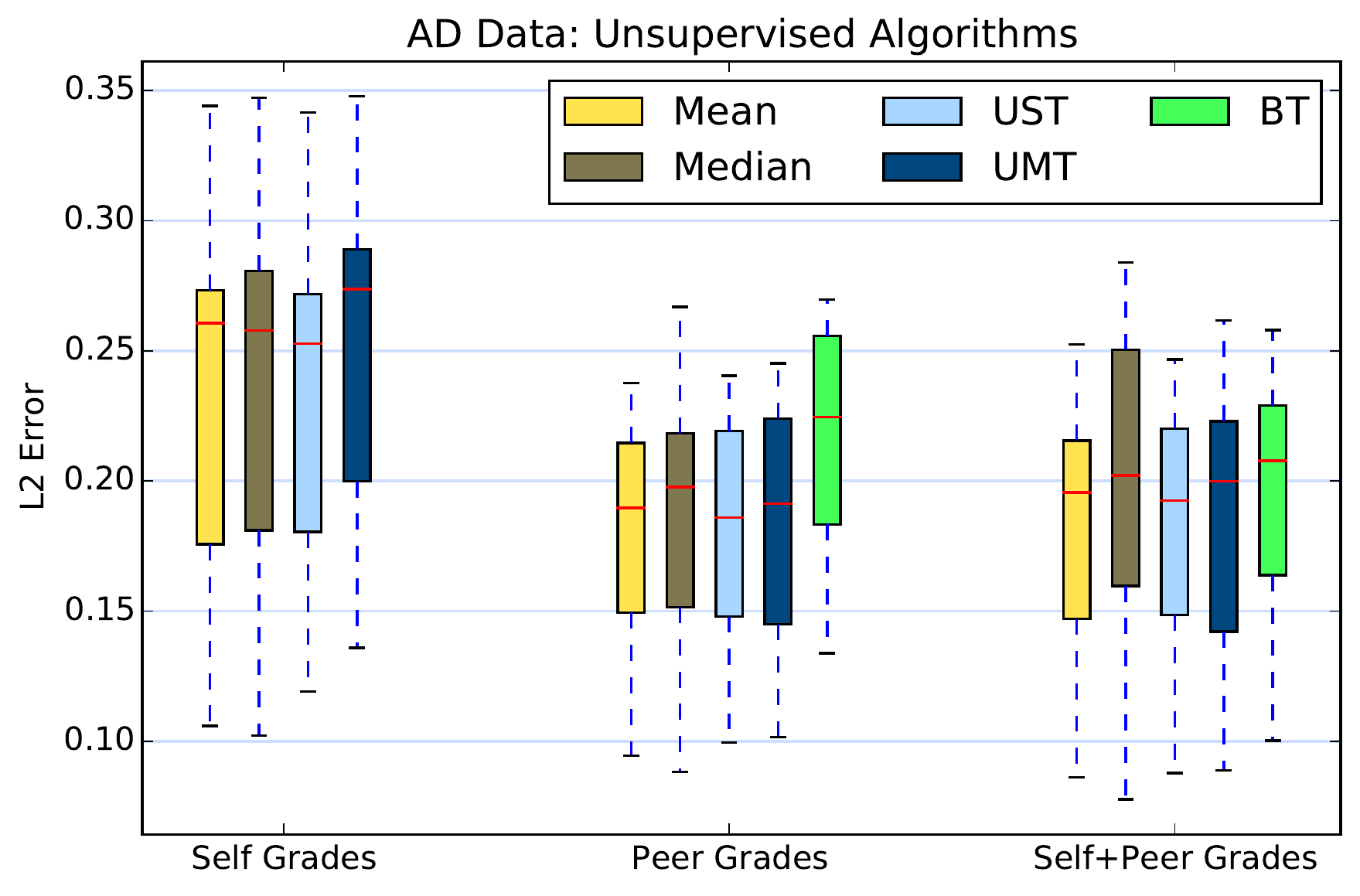}
\hfill 
\includegraphics[width=0.49\textwidth]{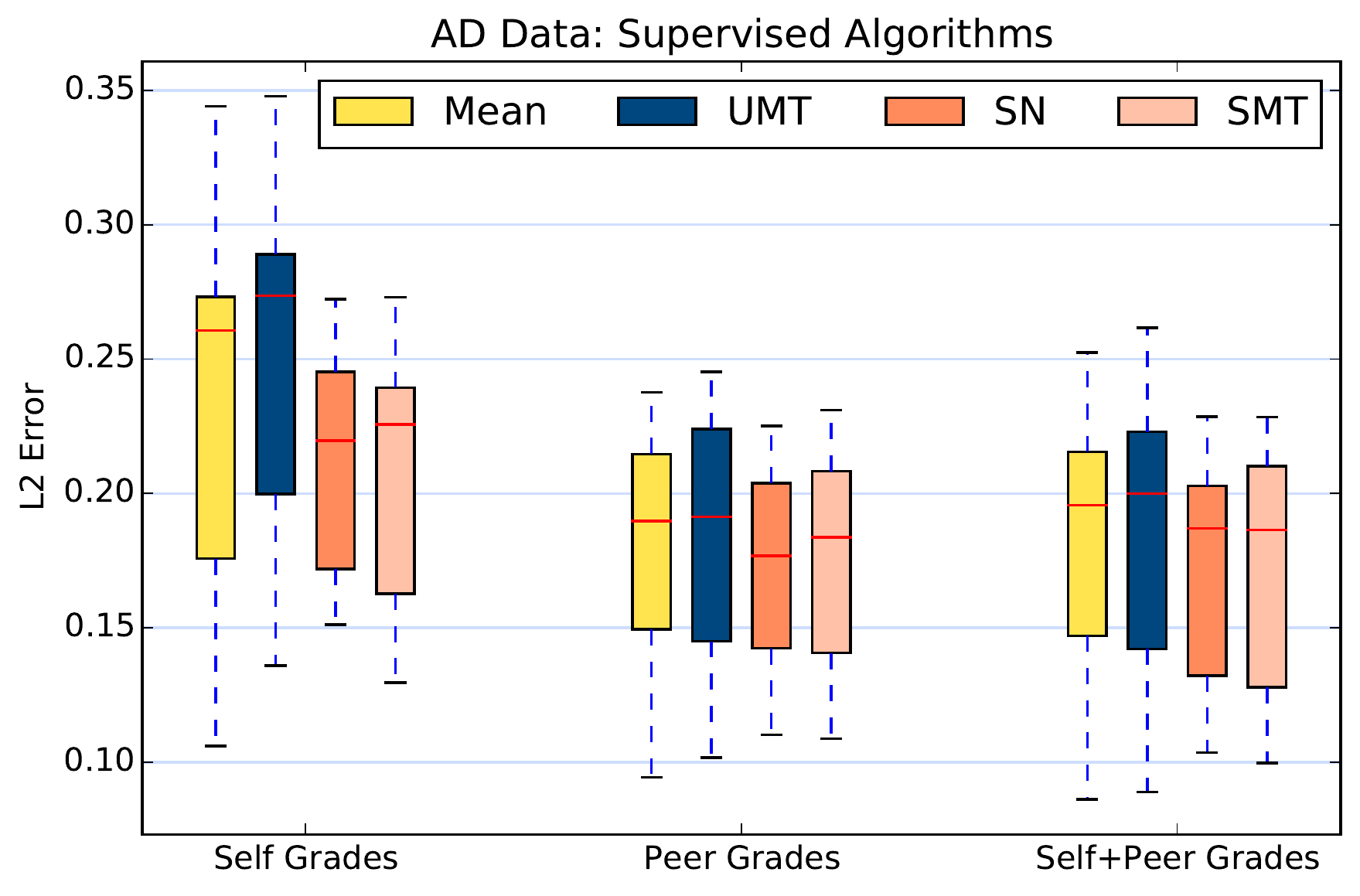}
\includegraphics[width=0.49\textwidth]{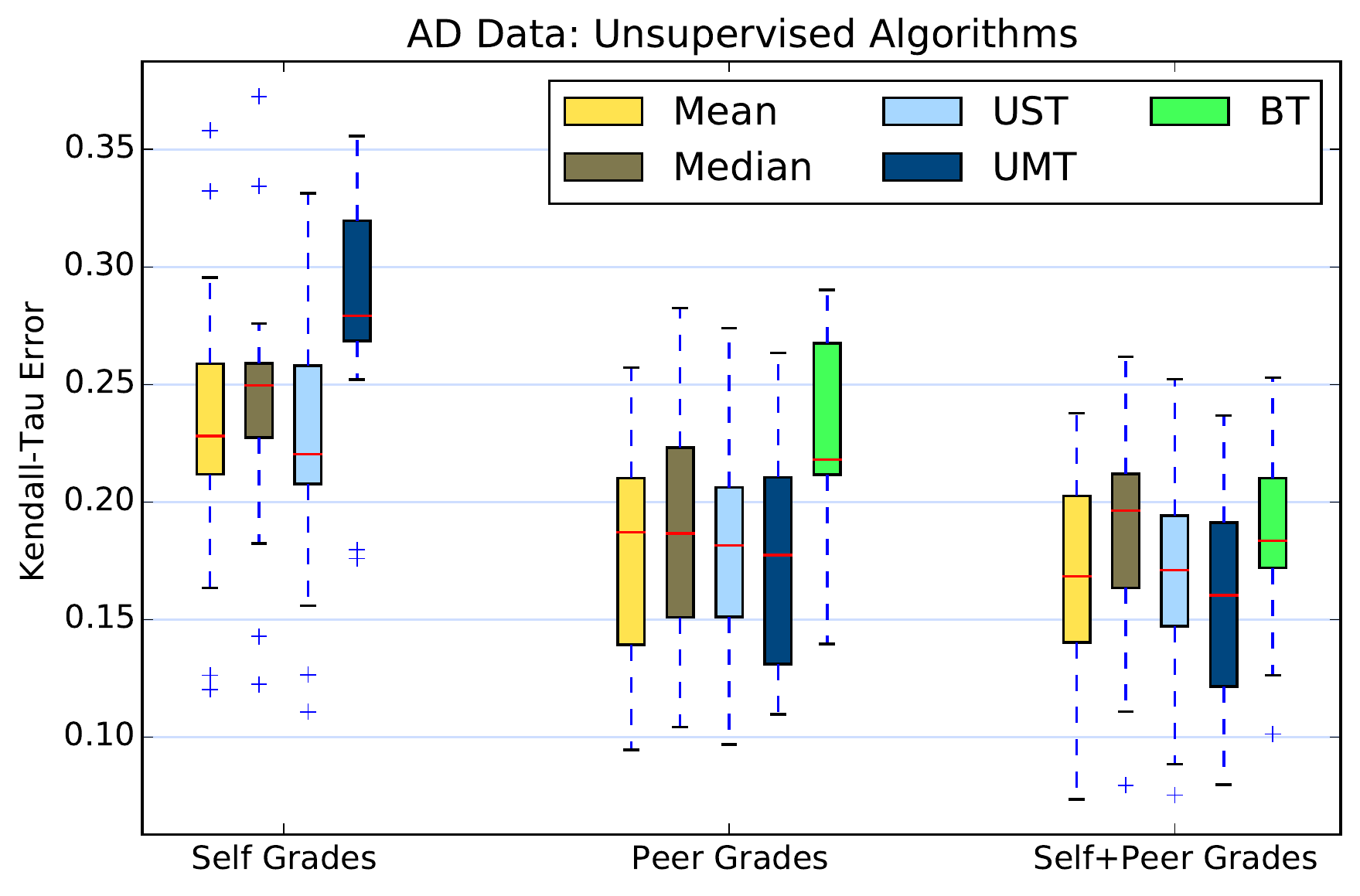}
\hfill 
\includegraphics[width=0.49\textwidth]{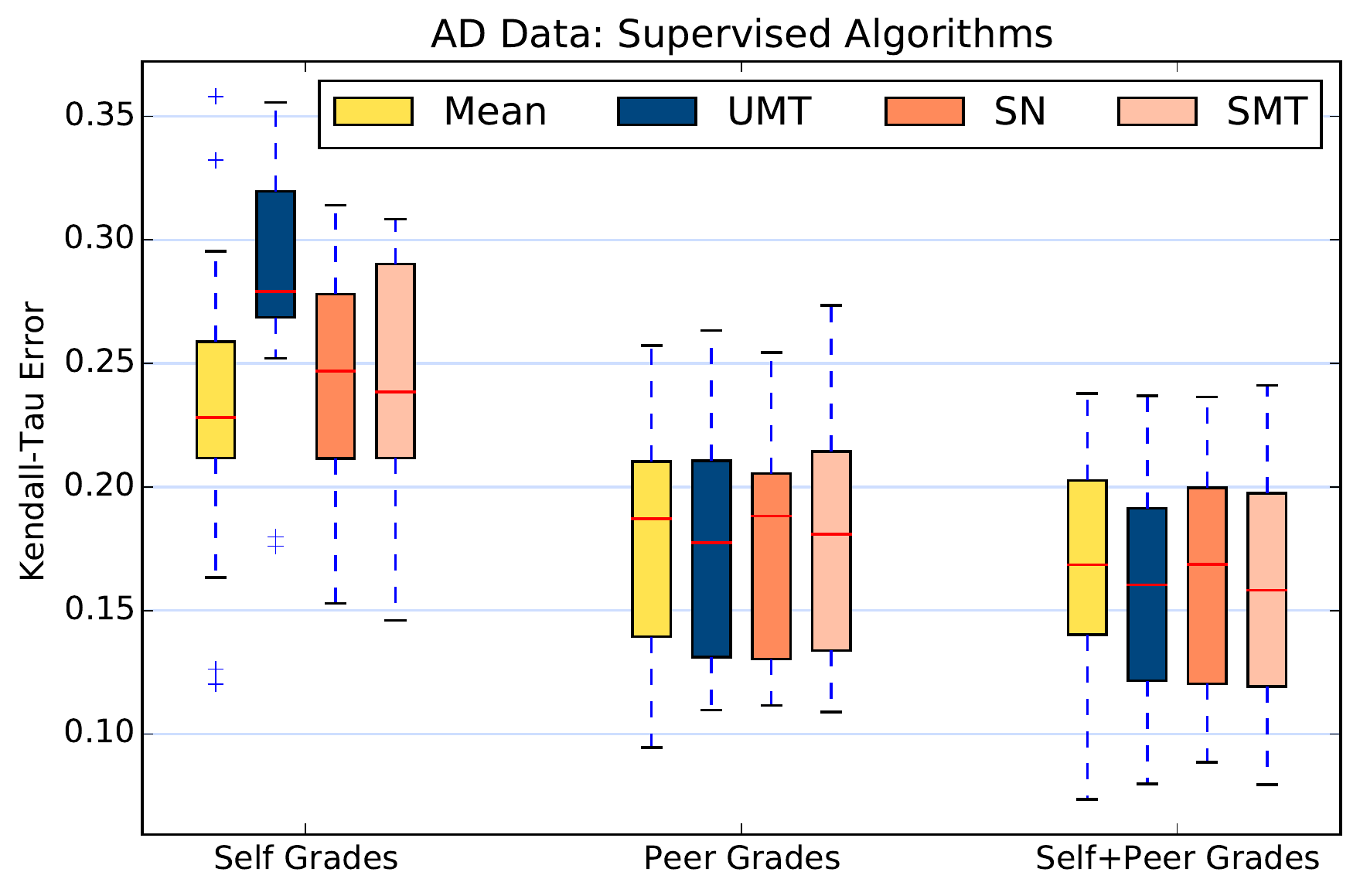}
\caption{
Fitting cardinal models to our AD data. The box plots show the $L_2$ errors (top) and Kendall-$\tau$ errors (bottom) with respect to the TA grades of each exercise over the whole semester. In each panel, the three groups refer to the data that was used to fit the models: the 3 self grades only, the 6 peer grades, or both together. }
\label{fig-results-summary}
\end{figure*}

A second aspect illustrated in Figure~\ref{fig-histogram-self-peer-ta} (right)
visualizes the relationship between TA grades and peer
grades for individual submissions. While the tendency of peer grades generally being higher than the TA grades
is again evident here, we can also see quite a number of submissions that received a score of 0 by
TAs but moderate to large scores by peers. Looking into the
corresponding submissions reveals that a typical reason is a wrong solution to the exercise where many reviewers miss the error and give full marks instead.
We can also see in the figure that there is a large number of
submissions with full TA grades but slightly lower peer grades. This happens because it is simply unlikely for all 6 peer grades to be full scores and a single lower grade will drag down the mean value.

\begin{figure*}[tb]
\includegraphics[width=0.475\textwidth]{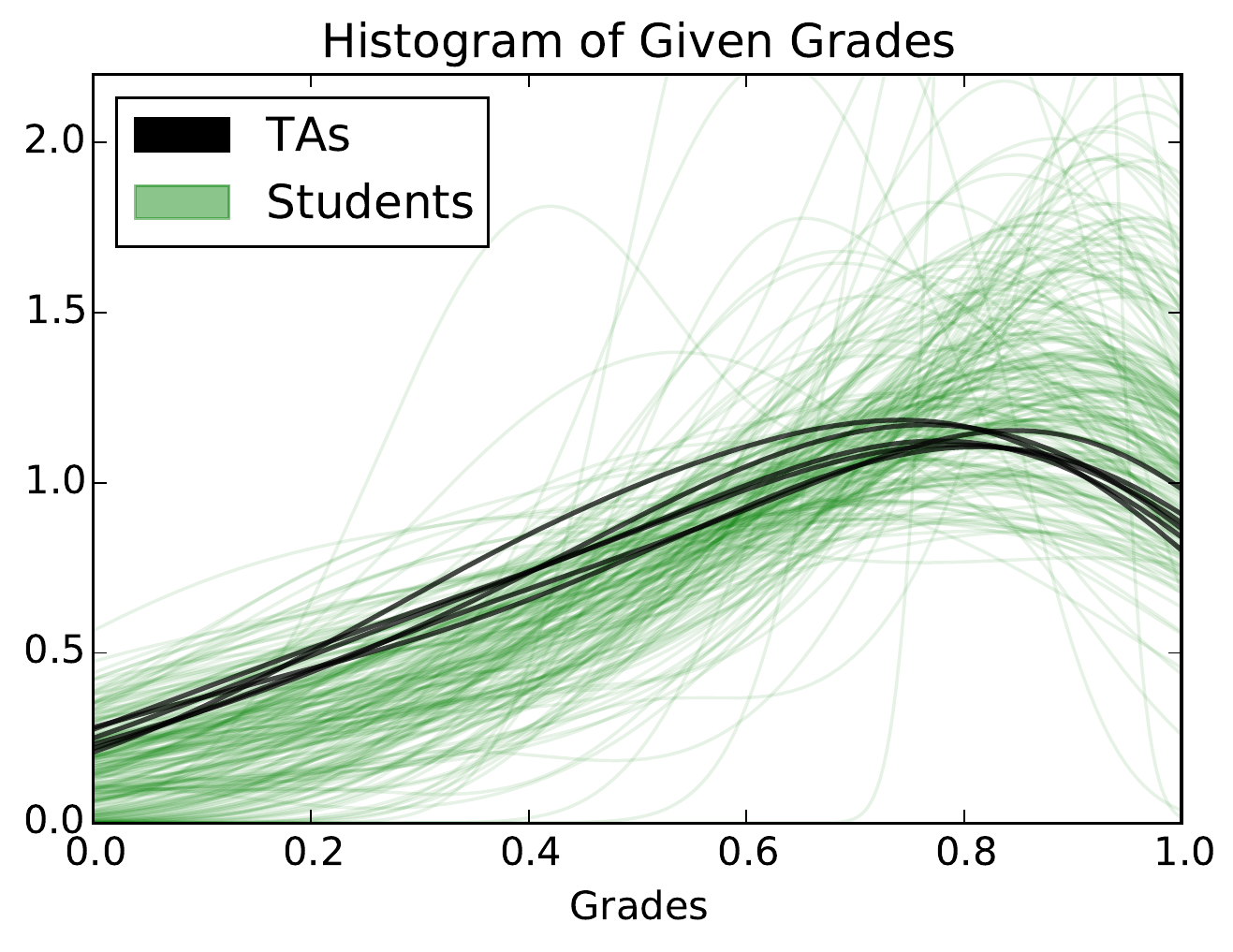}
\hfill 
\includegraphics[width=0.505\textwidth]{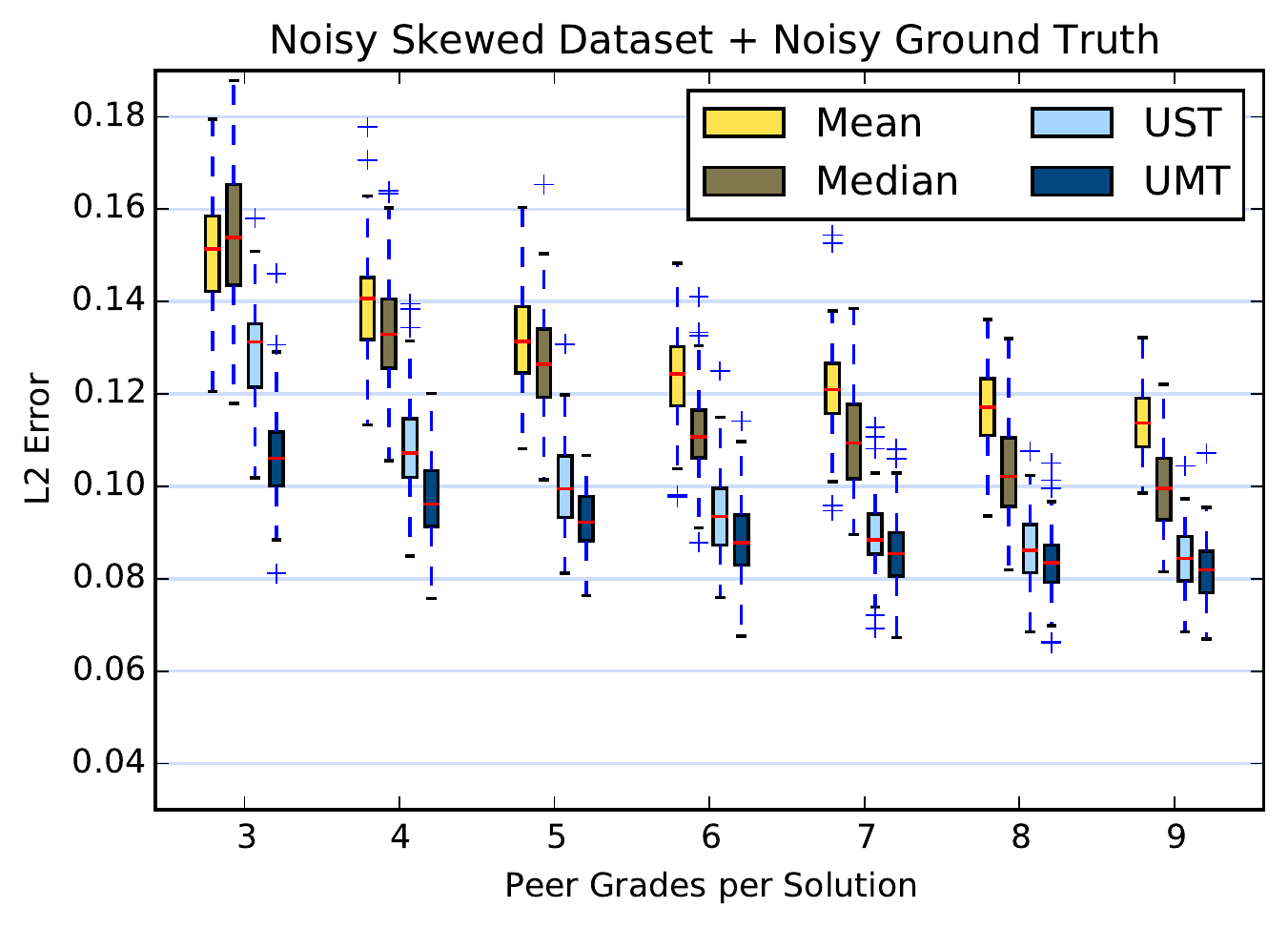}
\caption{
Analyzing the effect of using several TAs as baseline. Left: Smoothed
histograms of all given grades by the TAs and students. Right: Box plots in the same setting as in
Figure~\ref{fig-results-summary} (top right), but with white noise
added for the evaluation step to simulate the effect of using several
TAs: errors were computed against noisy true grades (independent
Gaussian noise $\mathcal N(0, 0.05^2)$).}
\label{fig-gradehistograms}
\end{figure*}

On a more abstract level, the figure reveals that 
the ``sources of error'' for a grader are not only a bias
due to different taste or different levels of strictness as suggested
in the probabilistic models, but also a
serious lack of understanding or information. This is problematic
for statistical algorithms: if a submission gets
high scores by most peer graders, there is no way to know 
whether this is really justified because the given solution is
correct or whether the reason is that all peer graders have overlooked a
crucial mistake in the submission. On the other hand, it is hard to detect cases of otherwise reliable graders getting a score completely wrong, in particular in the realistic scenario where only few grades are given for each submission.

\subsection{Fitting the models to the data}
We analyze our data using the introduced models.
In the unsupervised scenario, we simply take all grades,
fit the models, and estimate a ``true score''. Note that in
the unsupervised setting, we cannot optimize the model to fit any
ground truth (such as the TA grades). If the histogram of peer grades
is shifted with respect to the TA grades\footnote{See e.g.\ in the third panel
of Figure~\ref{fig-histogram-self-peer-ta} (left).}, there is no way for an
unsupervised model to correct for this.
Hence, comparing the estimated true scores to TA scores by any loss function
that compares the scores directly, such as the $L_2$ error, might be
dominated by the overall bias shift. To cover for this, we use the
Kendall-$\tau$ rank correlation as a second error measure.

The hyperparameters of the models are chosen as follows.  The
parameters $\mu_{score}$, $\sigma_{score}^2$ and $\sigma_{bias}^2$ for
UST and UMT only control the strength of regularization and were found
to have little to no impact on the overall accuracy. The reliability
parameters $\alpha$ and $\beta$ control whether the model gives all
students a similar reliability or fits them to a large variety of
reliability values. In our evaluations, we use the sample mean and
variance of the given peer grades for each exercise to set
$\mu_{score}$ and $\sigma_{score}^2$ and fix the remaining
hyperparameters at $\sigma_{bias}^2=\nicefrac{1}{36}$, $\alpha=3$ and
$\beta=\nicefrac{1}{30}$. For BT, we again use the sample mean and
variance as priors and choose $\alpha=10$, $\beta=2$ for the
reliability. Note that in the exercise sheet where we applied the pure
ordinal setting, students only report rankings of submissions, so sample
mean and variance are unknown there. In that case, the parameters can be
chosen arbitrarily to control the distribution of the resulting scores.

\subsection{Analysis of unsupervised models}
We will begin with a comparison of the \textbf{ordinal algorithms} (For this section, the implementation provided by \cite{RamJoa14} was used.) A plot of the algorithm performance on each individual exercise is shown in Figure~\ref{fig-ordinal-results}. In our case study we usually collected numeric grades -- only their induced rankings are fed into the ordinal algorithms. For that reason, we again use the Mean algorithm as baseline. Exercises 51 and 52 are an exception to that, as we collected the grades in an ordinal fashion there, so we use the simple BC algorithm as baseline for those two exercises.
With the exception of Mallow's model, all ordinal algorithms perform very similarly. In particular, using only the peer grades they all perform worse than the mean algorithm by a significant margin. The difference is smaller when the peer and self grades are combined, as there are significantly more pairwise comparisons in this dataset which helps the ordinal models. Adding the reliability estimation to the models curiously increased the Kendall-$\tau$ errors in most exercises by 0.01-0.03 (not shown in the figure). On the exercises 61 and 62, we asked each student to grade 5 other submissions instead of 2. The effect is visible on the peer grades, as the performance of the ordinal algorithms is substantially better on those tasks, strengthening the point that ordinal models inherently need more grades to match the performance of cardinal algorithms.

As a further experiment, we collected the grades in an ordinal fashion in exercises 51 and 52. Much to our surprise, the embarrassingly simple approach of the BC algorithm yields no worse results than the rankings of the more complex ordinal algorithms on exercise 51 and even beats them on exercise 52. This could imply that more complicated models are not needed for ordinal peer grading.
Additionally, we can see that the performance of the ordinal algorithms on exercises 51 and 52 is rather bad in general. One reason for this might be the grading behavior of the students in ordinal tasks. Many students reported that they made less of an effort for ordinal peer grading than for cardinal peer grading because they considered it easier to quickly come up with a ranking than to report absolute grades. We believe that this might be a serious drawback of collecting peer grades in an ordinal fashion. If students do not look at the details of a submission, it is unlikely that the overall grading performance improves over cardinal grading.

Figure~\ref{fig-results-summary} shows the results of the \textbf{cardinal models} on our AD dataset along with BT representing the ordinal models. Note that BT cannot be run exclusively on the self assessment grades, as each student only graded one submission which does not imply any ranking. As opposed to the results on the artificial data in Figure~\ref{fig-artificial} where the model-based approaches clearly outperform the baseline, we can now see that UST and UMT provide no improvement over the simple mean. This finding is disappointing and contradictory to the results in the literature. We will discuss possible explanations for this behavior in the following section.

\subsection{Why do the models provide no improvements over mean?}

\textbf{Amount of data.} We have about 6 peer grades and 3 self grades per submission, collected over 19 exercises. Each student submitted around 57 grades in total. The results on artificial datasets as well as other studies on peer grading suggest that this should be enough to get a reasonably reliable estimate. A lack of data is not the problem here.

\textbf{Model assumptions mismatch.} As seen above,
  our data usually does not satisfy the
  model assumptions (normal distributions, etc). However, experiments with
  artificial data whose distributions do not agree with the model
  assumptions show that the model typically still works reasonably well. We
  do not believe that the model mismatch is the major source of the
  problem.

\textbf{Model fitting.} We set the hyperparameters as described above, though in our experiments, we found that the models are not sensitive to the choice of the hyperparameters. The actual model fitting was done with the EM algorithm. \cite{PiechEtal13} reported that the results of the EM algorithm are almost equal to the ones of Gibbs sampling. Furthermore, simulations with a number of different artificial datasets resulted in consistent estimates for the score, bias and reliability values.

\textbf{TAs as baseline.} We evaluate our errors against the TA
  grades, that is we consider the scores or orderings given by the TA
  grades as ground truth. However, these grades have been given by 6
  different TAs, so the variance within these grades might make them
  unsuitable to serve as ground truth (in the extreme case, if we
  compared against random grades, then none of the models would
  outperform the others). We first study this effect with artificial
  experiments. Consider the setting in
  Figure~\ref{fig-artificial}, but we now add noise to the true
  grades before computing the errors (we use Gaussian noise with
  standard deviation $0.05$).
The results can be seen in
  Figure~\ref{fig-gradehistograms} (right). They still look
  similar to the original results in Figure~\ref{fig-artificial}, just
  the overall performance worsened slightly due to the noise. In
  particular, the UST and UMT models still considerably outperform the
  mean estimate. This is even the case if we use an unrealistically large
standard deviation for the noise, say $0.2$.

\begin{figure*}[tb]
\includegraphics[width=0.33\textwidth]{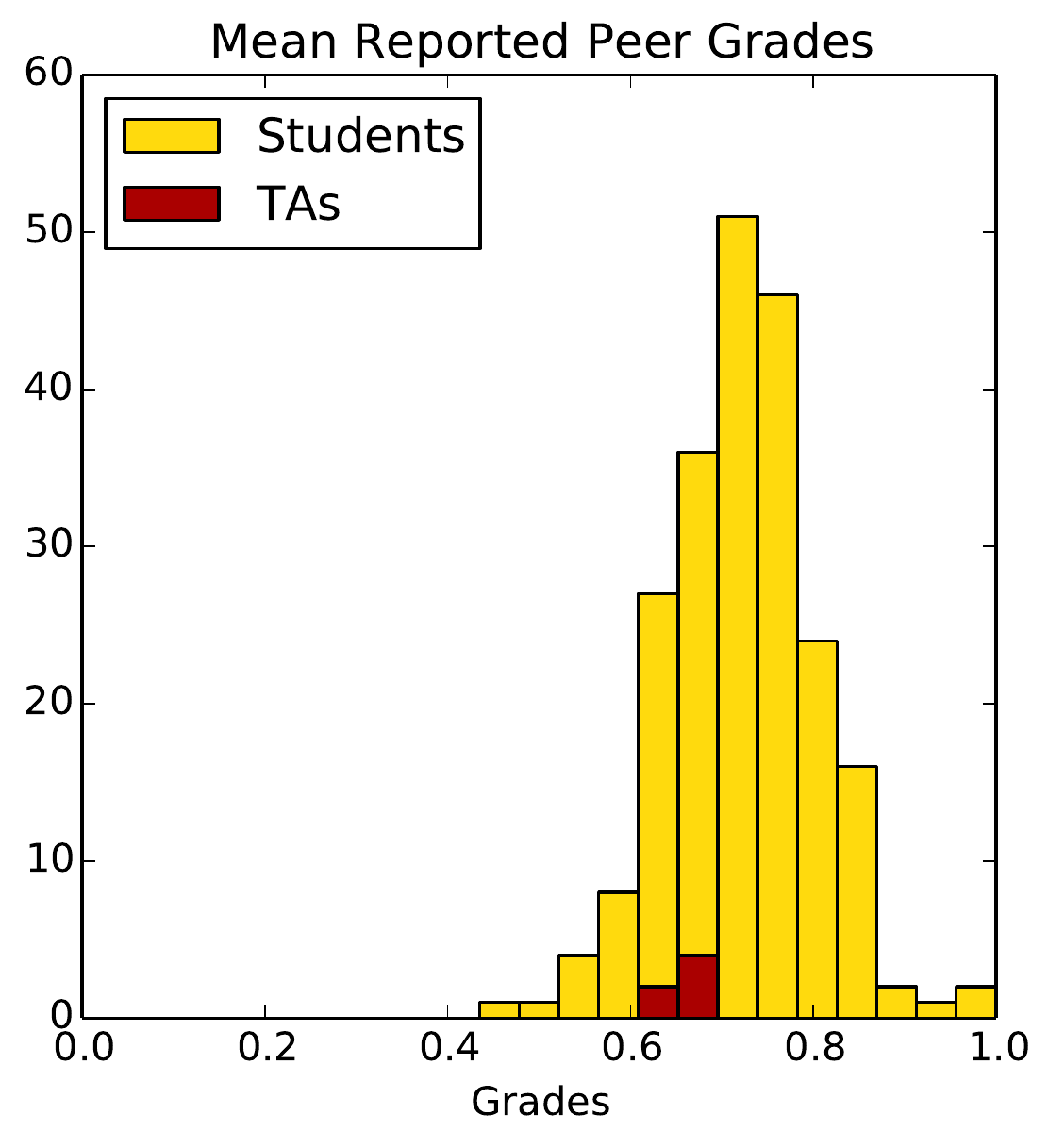}
\includegraphics[width=0.33\textwidth]{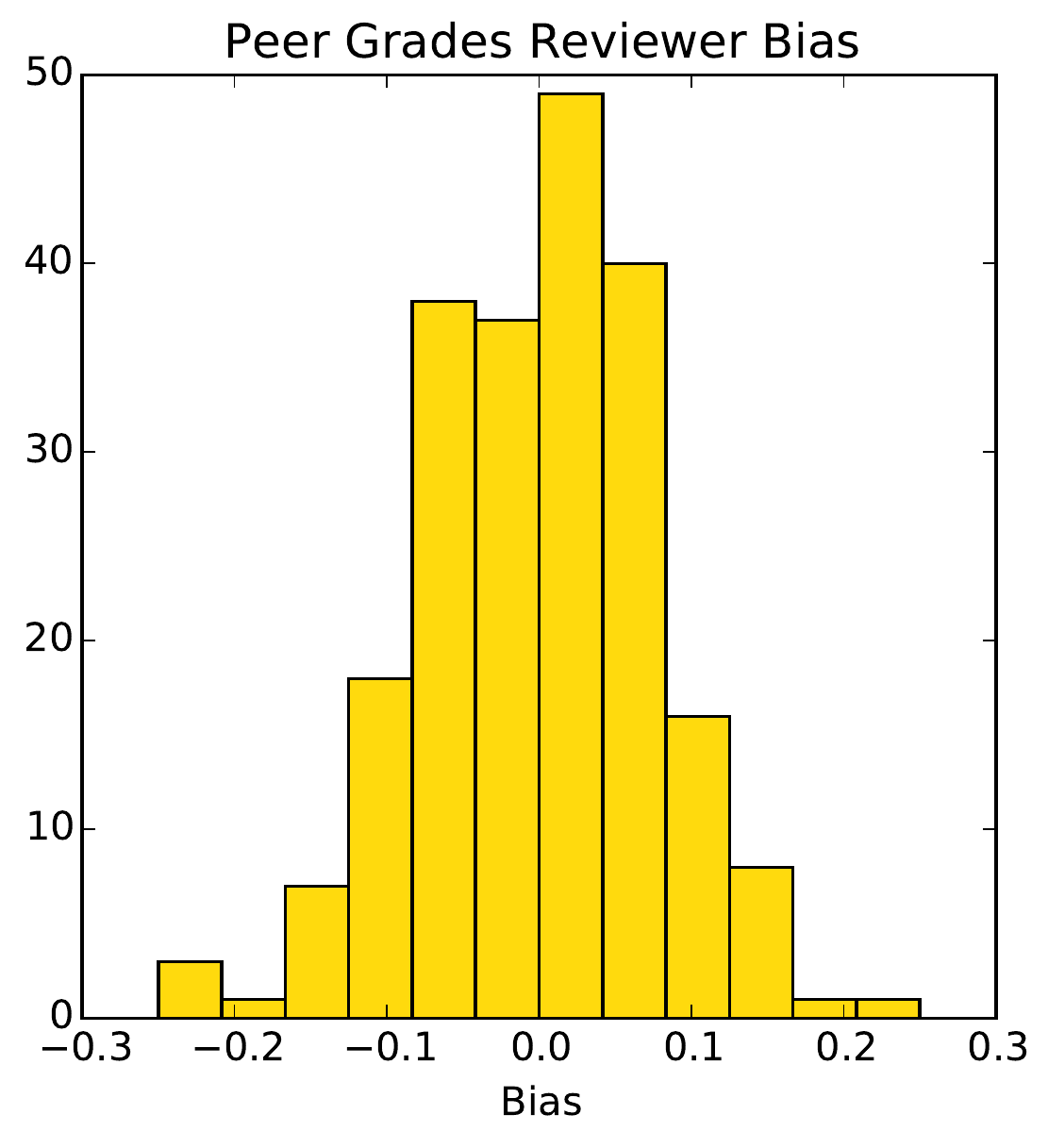}
\includegraphics[width=0.322\textwidth]{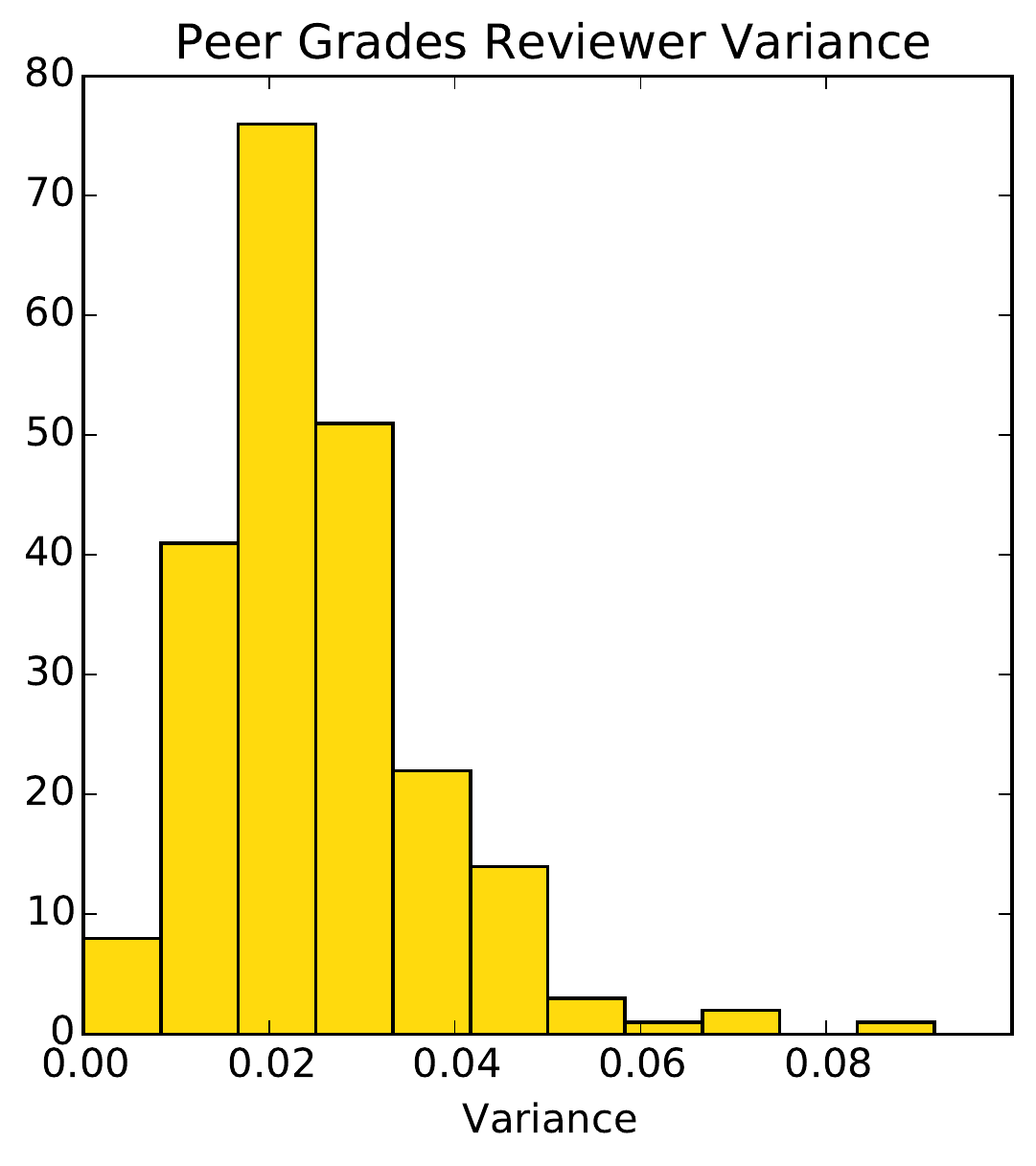}
\caption{Left: Histogram depicting the mean of all peer grades that each student (and TA) reported over the course of the semester. Mid, right: For each student and all exercises, the deviations of the student's reported grade from the mean of all other grades for the same submissions are calculated. The sample mean of those deviations is the reviewer bias while the sample variance is similar to the inverse of the reliability in the UST and UMT models.}
\label{fig-ta-peer-biases}
\end{figure*}

As a next step, we look at our actual data to evaluate
  the consistency among the TA grades in comparison to the
  consistency among the peer grades. Due to constraints during data collection, we did not have the possibility to conduct an extensive experiment to compare the grading performance of the TAs with each other. Instead, we consider the following evaluations. We
  first compare the average reported grade of the TAs to the average
  grades given by the student reviewers. As can be seen in
  Figure~\ref{fig-ta-peer-biases} (left), the TAs have a very low bias
  amongst each other, in particular it is much lower than the biases
  amongst the students. Next, we look at the overall histograms of all
  given grades by each TA, see Figure~\ref{fig-gradehistograms}
  (left). There is little variance amongst the TA histograms, in
  particular compared to the variance in the peer histograms. All
  in all it looks like the TAs grade reasonably
  consistently, so we believe that the use of different TA
  grades as ground truth cannot be the major reason for the lack of
  improvement of the probabilistic models over the simple baseline.

 \textbf{Bias vs. reliability.} As seen in
  Figure~\ref{fig-ta-peer-biases} (mid, right), the overall biases are
  not very large, only few exceed an absolute value of {|0.1|}. On the other hand, the variance in reporting grades is quite
  high. More than half the students reported grades that deviate from
  the mean peer grades with a standard deviation of at least {0.14} (a
  variance of roughly {0.02}). To gain some intuition on the values,
  note that a student who theoretically always reported the same
  score {0.72} for all submissions (the overall mean peer grade on all
  exercises together) would end up with a variance of {0.05}. 

 It was reported by \cite{PiechEtal13} that more than 90\% of UST's improvements are due to the fact that the  model-based approaches correct for the bias, an effect that we also confirmed on artificial data. 
However, in our AD data, the 
  errors induced by low reliabilities dominate the
  errors due to bias. This may be part of the reason why the models do
  not perform well here.

\textbf{Sources of grading error.} As discussed above, reasons for
  differences in grading behavior are not only that people have
  ``different tastes'' or are ``differently strict''. Rather, it is
  often the case that graders make serious errors due to lack of
  information or lack of understanding in the topic. This
  problem is bound to come up when using peer grading in a course such
  as algorithms and data structures and might be much less of an issue
  when grading is used to evaluate project reports \cite{RamJoa14} or
   design questions \cite{KulkarniEtal15}.

To check whether the model-based algorithms improve if we just use ``easy-to-grade''
exercises, we selected a number of exercises where the errors were low, indicating that the students
had no difficulties in grading the submissions.
We found that even in this scenario,
the model-based algorithms
do not  perform better than the mean. This may be due to the fact that
in easy-to-grade exercises, the mean algorithm does a good enough job
at eliminating the different biases or reliabilities, so there is not 
much room left for improvements. Similarly, if we just train on
``difficult-to-grade'' exercises, we do not find an improvement of
model-based algorithms compared to the mean. The reason is that in
this scenario, 
the errors are dominated by the submissions that are graded totally
wrong by a large number of students, so none of the algorithms can do a good job.

 \textbf{Unmotivated students.} One might suspect that some
  proportion of students tried to get away with minimal effort and
  produced grades that are pretty much random. However, looking
  closely into the data reveals that we only had a very small number of
  these students, so the models should succeed at giving them low
  reliabilities and lead to a better performance.

\subsection{Analysis of supervised models}

Considering the results for the supervised learning models in
Figure~\ref{fig-results-summary}, we see a similarly disappointing picture: the
supervised models do not improve over the simple mean
estimator. While the Kendall-$\tau$ errors do not
change much compared to the unsupervised models, the variance in $L_2$ errors over different tasks gets smaller. This is an effect of calculating the student reviewer bias values against the TA grades which improves the overall error in tasks that have a very high error and overfits in tasks that were graded well by most students in the first place. All in all, supervision does not seem to have a significantly positive effect on the performance of our algorithms.

\subsection{Analysis of further correlations in the data}

To see whether we can meaningfully improve the performance of the models, we first check the correlation between the mean homework performance and the overall bias or mean deviation of the students' reported grades compared to the TA grades. For bias, we use the same formula as in the SN model. The mean deviation for a student $g$ is likewise calculated on all reported grades $s_1,\ldots,s_k$ over the whole course in relation to the TA grades $\score(s,true)$ as
$$\frac{1}{k} \sum_{i=1}^k \left|(\score(s_i,g) - \bias(g)) - \score(s_i,true)\right|.$$

The homework performance and the bias of the students are only weakly correlated with an $r$-value\footnote{Pearson product-moment correlation coefficient.} of $-0.22$, i.e.\ students with higher grades on their own submissions tend to report lower grades. However, this is not due to better students being stricter or having higher standards but rather because better students are more likely to find flaws in the submissions and therefore report lower grades on average. While the former would necessitate a correction, this would punish students for doing a good job at finding large mistakes and results in a worse overall performance concerning grade accuracy.

The correlation between the mean deviation and the homework grades is even weaker at an $r$-value of only $-0.10$, indicating a marginally higher agreement between well performing students and the TAs. A possible explanation for the low correlations is the fact that students handed in submissions in groups which means that the scores of each student do not necessarily reflect their knowledge on the topic. A more personal measure is given by the exam that was written at the end of the course. First, we note that the exam grades and the mean homework grade for the students are correlated with an $r$-value of $0.28$, a rather low value. Taking the exam grades instead of the mean homework grade, the correlation with the bias is now slightly higher at $r=-0.26$ while the mean deviation is correlated with the exam grades at $r=-0.28$ which is much stronger than the correlation with the homework scores.

We incorporate the exam grades as a measure of reliability into the model using both mentioned approaches. To use the exam grades directly as the (constant) reliability for each student in the UMT model, we simply scale them to lie in the range $[0,150]$, resulting in a distribution similar to the Gamma-distribution for the estimated reliability in UMT. For the second approach, we estimate the reliability as usual in UMT but then multiply these values with the normalized exam grades (exam grade divided by mean exam grade). Both approaches yield similar results that are only marginally better than UMT. It seems that using the exam grade for the reliability compensates for some of the error that UMT introduces but still fails to significantly improve over a simple mean.

\section{Conclusions}

We have mixed feelings towards peer grading after
evaluating all the data collected in our algorithms and data
structures course.
The positive point of view is that even though peer grades tend to
be more optimistic than TA grades, the size of this effect is not
very large. The peer grades give a reasonably informed picture of the
true grades. For this reason, using simple estimates such as the mean
grade is competitive to more elaborate model-based algorithms. From
an application point of view this finding is helpful: 
an easy to understand mechanism such as a simple mean is more acceptable to students than a complicated model when it comes to generating their final grades.

From a statistical or machine learning point of view, our results are somewhat
disappointing. None of the models we tested outperforms the simple
mean estimator on our data. Our general feeling is
that it will be very difficult to come up with algorithms that do a
much better job in this particular setting. The reasons for this difficulty might
be the heterogeneity of the score distributions of different exercises, the
high variance among graders and the different and rather
unpredictable sources of grading errors (``lack of understanding''
rather than a ``slightly different taste''). It seems very hard to
model all these aspects unless one has a much larger amount of grades
per submission. However, the peer grading setup does not allow us to simply
scale up the number of grades per submission, because students are not
willing to invest even more time into this process. Finally, let us mention that our negative findings may be special to a course such as algorithms and data structures. Peer grading could be more successful for tasks where grading is a matter of taste rather than of understanding.

As opposed to other papers in the literature, we could not confirm the hypothesis that collecting grades in an ordinal rather than in a cardinal fashion leads to improved estimates. To the contrary, the students themselves reported that they tend to be more sloppy when being asked to provide ordinal grades, so we are somewhat pessimistic about ordinal grading in general.

The data set generated by our class as well as a moodle
plugin that supports peer grading in a group-based scenario
as used in the AD course have been been made publicly available
on our homepages.

\section*{Acknowledgments}
This work has been supported by the German Federal Ministry of Education and Research (01PL12033, Universitätskolleg/TP 16 Lehrlabor at Universität Hamburg), the German Research Foundation (LU1718/1) and the Institutional Strategy of the University of Tübingen (Deutsche Forschungsgemeinschaft, ZUK 63). The authors alone are responsible for the content of this publication.

%
%
%\bibliographystyle{SIGCHI-Reference-Format}
%\bibliography{peer_grading_literature}

\begin{thebibliography}{00}

%%% the Web address.

\ifx \showCODEN    \undefined \def \showCODEN     #1{\unskip}     \fi
\ifx \showDOI      \undefined \def \showDOI       #1{{\tt DOI:}\penalty0{#1}\ }
  \fi
\ifx \showISBNx    \undefined \def \showISBNx     #1{\unskip}     \fi
\ifx \showISBNxiii \undefined \def \showISBNxiii  #1{\unskip}     \fi
\ifx \showISSN     \undefined \def \showISSN      #1{\unskip}     \fi
\ifx \showLCCN     \undefined \def \showLCCN      #1{\unskip}     \fi
\ifx \shownote     \undefined \def \shownote      #1{#1}          \fi
\ifx \showarticletitle \undefined \def \showarticletitle #1{#1}   \fi
\ifx \showURL      \undefined \def \showURL       #1{#1}          \fi

\bibitem{Diez13}
{J. D{\i}ez}, {O. Luaces}, {A. Alonso-Betanzos}, {A. Troncoso}, {and} {A.
  Bahamonde}. 2013.
\newblock \showarticletitle{Peer assessment in {MOOC}s using preference
  learning via matrix factorization}. In {\em NIPS Workshop on Data Driven
  Education}.
\newblock


\bibitem{KulkarniEtal15}
{C. Kulkarni}, {K.~P. Wei}, {H. Le}, {D. Chia}, {K. Papadopoulos}, {J. Cheng},
  {D. Koller}, {and} {S.~R. Klemmer}. 2015.
\newblock \showarticletitle{Peer and self assessment in massive online
  classes}. In {\em Design Thinking Research}. 131--168.
\newblock


\bibitem{MiYeung15}
{F. Mi} {and} {D.-Y. Yeung}. 2015.
\newblock \showarticletitle{Probabilistic graphical models for boosting
  cardinal and ordinal peer grading in {MOOC}s}. In {\em 29th AAAI Conference
  on Artificial Intelligence}.
\newblock


\bibitem{PiechEtal13}
{C. Piech}, {J. Huang}, {Z. Chen}, {C. Do}, {A. Ng}, {and} {D. Koller}. 2013.
\newblock \showarticletitle{Tuned models of peer assessment in {MOOC}s}. In
  {\em International Conference on Educational Data Mining}.
\newblock


\bibitem{RamJoa14}
{K. Raman} {and} {T. Joachims}. 2014.
\newblock \showarticletitle{Methods for ordinal peer grading}. In {\em
  International Conference on Knowledge Discovery and Data Mining (SIGKDD)}.
  1037--1046.
\newblock


\bibitem{ShahEtal13}
{N.~B. Shah}, {J.~K. Bradley}, {A. Parekh}, {M. Wainwright}, {and} {K.
  Ramchandran}. 2013.
\newblock \showarticletitle{A case for ordinal peer-evaluation in {MOOC}s}. In
  {\em NIPS Workshop on Data Driven Education}.
\newblock


\bibitem{Topping98}
{K. Topping}. 1998.
\newblock \showarticletitle{Peer assessment between students in colleges and
  universities}. In {\em Review of Educational Research}, Vol.~68. 249--276.
\newblock


\bibitem{Venanzi14}
{M. Venanzi}, {J. Guiver}, {G. Kazai}, {P. Kohli}, {and} {M. Shokouhi}. 2014.
\newblock \showarticletitle{Community-based bayesian aggregation models for
  crowdsourcing}. In {\em Proceedings of the 23rd International Conference on
  World Wide Web}. 155--164.
\newblock


\bibitem{VozHolGil14}
{A. Vozniuk}, {A. Holzer}, {and} {D. Gillet}. 2014.
\newblock \showarticletitle{Peer assessment based on ratings in a social media
  course}. In {\em International Conference on Learning Analytics And
  Knowledge}. 133--137.
\newblock


\bibitem{Walsh14}
{T. Walsh}. 2014.
\newblock \showarticletitle{The {PeerRank} method for peer assessment}. In {\em
  European Conference on Artificial Intelligence}.
\newblock


\end{thebibliography}

\end{document}